\documentclass{article}

\usepackage{arxiv}

\usepackage[T1]{fontenc}
\usepackage[babel=true]{csquotes} 
\usepackage[fleqn]{amsmath} 
\usepackage{amsthm} 
\usepackage{booktabs} 
\usepackage{multirow} 
\usepackage{amssymb} 
\usepackage{float}
\usepackage{hyperref} 
\usepackage[french]{cleveref} 
\usepackage{algorithm}
\usepackage[noend]{algpseudocode}
\usepackage{mathrsfs}
\usepackage{mathtools}
\DeclarePairedDelimiter{\ceil}{\lceil}{\rceil}
\usepackage{graphicx}
\usepackage{pdflscape,array,booktabs}
\usepackage{threeparttable,booktabs,multirow,lscape}
\usepackage{subfig}
\usepackage{natbib}

\title{Ensemble Pruning via Margin Maximization}

\author{
  Waldyn Martinez \\
  Department of Information Systems \& Analytics\\
  Miami University\\
  Oxford, OH, 45056 \\
  \texttt{martinwg@miamioh.edu} \\
}

\begin{document}
\maketitle

\begin{abstract}
Ensemble models refer to methods that combine a typically large number of classifiers into a compound prediction. The output of an ensemble method is the result of fitting a base-learning algorithm to a given data set, and obtaining diverse answers by reweighting the observations or by resampling them using a given probabilistic selection. A key challenge of using ensembles in large-scale multidimensional data lies in the complexity and the computational burden associated with them. The models created by ensembles are often difficult, if not impossible, to interpret and their implementation requires more computational power than single classifiers. Recent research effort in the field has concentrated in reducing ensemble size, while maintaining their predictive accuracy. We propose a method to prune an ensemble solution by optimizing its margin distribution, while increasing its diversity. The proposed algorithm results in an ensemble that uses only a fraction of the original classifiers, with improved or similar generalization performance. We analyze and test our method on both synthetic and real data sets. The simulations show that the proposed method compares favorably to the original ensemble solutions and to other existing ensemble pruning methodologies.
\end{abstract}


\section{Introduction}
\label{intro}
Ensemble methods combine a large number of fitted values (sometimes in the hundreds) into a bundled prediction. The output of an ensemble method is generally the combination of many fits of the same data set by either reweighting the observations or by using subsets of the original set obtained from bootstrapping, resampling or other probabilistic selections of the data. There is sufficient empirical evidence pointing to ensemble performance being generally superior to that of individual or single classifiers \citep{ drucker1994boosting, breiman1996bagging, breiman1996bias, quinlan1996bagging, schapire98, opitz1999popular, dietterich2000ensemble, breiman2001random, maclin2011popular}. Boosting \citep{schapire1990strength} is one of the most well-known ensemble methods. The term boosting refers to a family of methods that combine weak classifiers (classification algorithms that perform at least slightly better than random) into a strong performing ensemble through weighted voting. AdaBoost \citep{freund97} is the leading implementation of boosting algorithms. 

Among ensembles, Bagging \citep{breiman1996bagging}, random forests \citep{breiman2001random} and rotation forests \citep{rodriguez2006rotation} are also strong performers in terms of their generalization ability. In addition to their ability to outperform individual classifiers, ensembles can also be very robust to overfitting, even when performing a large number of iterations \citep{quinlan1996bagging, schapire98}. To explain the successful performance of ensembles, \cite{breiman1999prediction} suggested that boosting, bagging and random Forests (which he referred to as arcing classifiers) reduce the variance, in the bias-variance decomposition framework, however \cite{schapire98} refuted this claim by empirically providing evidence that AdaBoost mainly reduced the bias. More importantly, \cite{schapire98} showed that AdaBoost is especially effective at increasing the margins of the training data. \cite{schapire98} developed an upper bound on the generalization error of any ensemble, based on the margins of the training data, from which it was concluded that larger margins should lead to lower generalization error, everything else being equal (sometimes referred to as the ``large margins theory"). The large margins theory has its roots in the margin separation framework in support vector machines (SVM) \citep{cortes1995support}. 

The proliferation of large scale, high velocity data sets, often containing variables of different data types, creates challenges for most traditional statistical and machine learning classification techniques, but it does so, even more markedly, for ensembles.  The term ``big data" has been used to describe large, diverse and complex data sets generated from various sources. The volume, variety and velocity (known as the 3Vs) are the main characteristics to distinguish big data problems from others \citep{megahed2013statistical}. A key drawback of fitting ensembles to large scale multidimensional data (big data) is their computational burden. The iterative nature of ensembles, as well as how complex the resulting solutions are, makes their implementation especially challenging. In addition, interpretations of ensemble predictions are not as straightforward as those of single classifiers and the implementation of the resulting models requires fitting the data through all of the iterations (sometimes in the hundreds) of the ensemble. A high number of iterations is oftentimes necessary to reap the benefits of the improved generalization performance provided by ensembles \citep{schapire98, dietterich2000ensemble}. For this reason, recent research effort has concentrated in reducing ensemble sizes, also called ensemble pruning (thinning), while trying to maintain or improve their predictive accuracy (see, e.g., \citealt{partalas2006ensemble, zhang2006ensemble, martinez2009analysis, chen2009predictive, tsoumakas2009ensemble, zhang2009multilayer, lu2010ensemble,   li2012diversity, dai2013competitive, germain2015risk}). There is also evidence that smaller ensembles perform as well as, or better than, their large counterparts (\citealt{zhou2002ensembling}), but knowing how large they should be is still an open research question. 

Ensemble pruning generally places additional computational costs on the training phase of the ensemble, due to the additional emphasis to identify a strong-performing subensemble, however a reduced ensemble translates into a more manageable and computationally less prohibitive model in the implementation phase \citep{chen2009predictive}.  Of particular importance in ensemble pruning is obtaining an ensemble that takes into account not only the quality of the individual classifiers, but also their disagreement \citep{zhang2009novel, germain2015risk}, that is, the effectiveness of ensembles depends also on the diversity of their componentwise classifiers, with the premise that more diverse classifiers perform better. Therefore, for high dimensional data sets, a more efficient algorithm could be constructed, if only the most diverse weak classifiers of the ensemble solution are taken into consideration in the final combination.
In this article we propose an algorithm that produces a reduced, strong-performing subensemble by optimizing the diversity of the resulting classifiers and maximizing its lower margin distribution. The proposed method is a weight-based quadratic optimization that aims to tune the weights of a given ensemble, such that the pairwise correlations of the weak classifiers and the margin variance are minimized, while the lower percentiles of the margin distribution of the ensemble are maximized.

\section{Preliminaries}
\label{prelim}

We assume a set of $T$ weak classifiers (also called weak learners), \(h_t(\textbf{x}), t = 1,2,...,T\), is created from the space (finite) of classifiers $\mathscr{H}$, each of which takes a \(p\times1\) input vector \(\textbf{x}\) and produces a prediction \(h_t(\textbf{x})\in \{-1,+1\}\) for a binary response variable \textit{Y}.  The combined classifier prediction of an ensemble with $T$ learners for a covariate vector \textbf{x} is given by:\\ 
\begin{equation}
\label{eq:1}
f_T(\textbf{x}) =\text{sign} \left(\sum_{t=1}^T \alpha_t h_t(\textbf{x})\right),
\end{equation}
where $\text{sign}:\mathbb{R}  \rightarrow \{-1, 0, +1\}$, such that $\text{sign}(a) = -1$, when $a <0$, $\text{sign}(a)=+1$ if $a >0$ and $\text{sign}(a)=0$ if $a=0$ (when $\text{sign}(a)=0$, we randomly assign $\{-1,+1\}$ to $f_T$); $\alpha_t$ is the weight associated with the $t^{th}$ weak learner, where $0\leq \alpha_t \leq 1$ and $\sum_{t=1}^T \alpha_t = 1$. The task of any ensemble or combined classifier $f$ is to create a set of weak learners and determine their associated weights $\{\alpha_1, \alpha_2, ..., \alpha_T\}$ based on a training sample of data pairs $S=\{( \textbf{x}_i, y_i ), i = 1,2, ...,n \}$ generated independently and identically distributed (i.i.d) according to an unknown joint distribution $P_{XY}$, to produce a combined prediction with small generalization (also called risk of the classifier)
\begin{equation}
\label{eq:2}
R[f] = E_{XY}g(y,f(\textbf{x})),
\end{equation}

\noindent for a given loss function $g$. For the binary classification framework, the generalization error is defined as $P_{XY}\left[f(\textbf{x}) \neq y\right]$, which is generally estimated by $\hat{P}_S \left[f_T(\textbf{x}) \neq y\right]  = \sum_{i=1}^n I(y_i \neq f_T(\textbf{x}_i))/n$, where $I(y_i \neq f_T(\textbf{x}_i)) = 1$ if $y_i \neq f_T(\textbf{x}_i)$, and 0 otherwise. We denote $P_{XY}[a]$ as the probability of event $a$ under the unknown distribution $P_{XY}$, and $\hat{P}_S[a]$ as the empirical probability of $a$ under $S$. We use $P[a]$ and $\hat{P}[a]$, when it is clear which distribution we are referring to. The weights assigned to the weak learners $\{\alpha_1, \alpha_2, ..., \alpha_T\}$ can be uniform, as in the case of random forests, or based the accuracy of the componentwise learners, as in the case of boosting. We will refer to the classifiers $h_t(\textbf{x})$ contained in an ensemble as weak learners, base learners or (individual) classifiers, and they are implemented by a base-learning algorithm $\textbf{B}$, that maps the input vector \textbf{x} to the binary response variable \textit{Y}. Base-learning algorithms can be decision trees, neural networks, or any other kind of learning or statistical method. The construction of an ensemble is based on two main steps, i.e., generating the weak learners, and then combining them. The final combination is done with a linear function, but the final prediction can also be based on user-specified thresholds.

To explain the, generally superior, performance of ensembles, \citet{schapire98} showed that margins were an integral part in understanding how ensembles could generalize. The margin of the $i^{th}$ training observation is defined by:

\begin{equation}
\label{eq:3}
m_i = m_i (\textbf{x}_i, y_i)= y_i \sum_{t=1}^T \alpha_{t} h_{t}(\textbf{x}_i).
\end{equation}

\noindent The margin can be viewed as a measure of ``confidence" of the prediction for the $i^{th}$ training observation and is equal to the difference in the weighted proportion of weak classifiers correctly predicting the $i^{th}$ observation and the weighted proportion of weak classifiers incorrectly predicting the $i^{th}$ observation, so that $-1\leq m_i \leq 1$. A margin value of $-1$ indicates that all of the weak learner predictions were incorrect, while a margin value of $+1$ indicates all of the weak learners correctly predicted the observation. Next, we will briefly define the most common ensembles to date: boosting \citep{schapire1990strength, freund97}, and random forests \citep{breiman2001random}. There are other ensemble methods that merit mention, but given the scope of this paper, we will only discuss these two.	

\subsection{Boosting Algorithms}

Boosting refers to the idea of converting a weak learning algorithm into a strong learner, that is, taking a classifier that performs slightly better than random chance and improving (boosting) it into a classifier with arbitrarily high accuracy. Boosting originated from the PAC (probably approximately correct) learning theory \citep{valiant1984theory} and the question that \citet{kearns1994cryptographic} posed on whether a “weak” learning algorithm can be boosted into an arbitrarily accurate ``strong" learner. AdaBoost, the most well-know boosting algorithm, has been shown to be a PAC (strong) learner. A strong PAC learner is formalized in the following definition:
\\

\noindent
\textbf{Definition 1} \citep{kearns1994cryptographic}. \textit{Let $\mathscr{F}$ be a class of concepts. For every distribution $P_{\textbf{X}Y}$, all concepts $f \in \mathscr{F}$  and all $\epsilon \in (0,1/2)$, $\delta \in(0,1/2)$, a strong PAC learner has the property that with probability at least $1-\delta$ the base learning algorithm $\textbf{B}$ outputs a hypothesis $h$ with $P\left[h(\textbf{x})\neq f(\textbf{x})\right] \leq \epsilon$. $\textbf{B}$ must run in polynomial time in $1/\epsilon$, and $1/\delta$ using only a polynomial (in $1/\epsilon$ and $1/\delta$) number of examples.}
\\

\noindent Boosting can be based on resampling or reweighting \citep{seiffert2008resampling}. The main goal of boosting methods is to give more voting power $\alpha_t$ to the weak learners or classifiers that perform the best. AdaBoost, for example, achieves this by iteratively using the same base-learning classifier, only modifying the weights of the observations $D_i^{t}$ at iteration $t$, therefore $\textbf{B}$ must accept observation weights $D_i$ as inputs. AdaBoost adaptively places more emphasis on the training observations that were misclassified by the previous weak learner (iteration). The weight that each observation receives in round $t+1$ of the iterations is given by

\begin{equation}
\label{eq:4}
D_i^{t+1}= \frac{D_i^t \exp⁡(-\alpha_t y_i  h_{t}(\textbf{x}_i))}{Z_t},
\end{equation}

\noindent where $Z_t = \sum_{i=1}^nD_i^{(t)}\exp\{-\alpha_ty_ih_{t}(\textbf{x}_i)\}$ is a normalization constant. The weights of misclassified observations increase by a factor of $\exp⁡(\alpha_t)$ at iteration $t$. The AdaBoost algorithm is described in Algorithm 1. The voting power of each base learner is given by $\alpha_t=  \frac{1}{2} \ln \left(\frac{1-\epsilon_t}{\epsilon_t}\right)$, where more emphasis is given to those base learners with lower misclassification error $\epsilon_t$. 

\begin{algorithm}
	\caption{AdaBoost}\label{AdaBoost}
	\label{alg:1}
	\begin{algorithmic}[1]
		\State $[Input]: S =\{(\textbf{x}_i,y_i), i= 1,...,n\}$ for $T$ iterations
		\State $[Initialize]: D_i^{(1)} = \frac{1}{n}$
		\State $[Loop]:$ do for $t = 1,...,T$
		\State \indent (a) obtain $h_t$ on the sample set $\{S, D_i^{(t)}\}$
		\State \indent (b) set $\epsilon_t = \sum_{i=1}^{n} D_i^{(t)}I \left(y_i \neq h_t(\textbf{x}_i)\right)$
		\State \indent (c) break if  $\epsilon_t = 0$ or $\epsilon_t \geq \frac{1}{2}$
		\State \indent (d) set  $\alpha_t = \frac{1}{2} \ln \left(\frac{1-\epsilon_t}{\epsilon_t}\right)$ 
		\State \indent (e) update  $D_i^{(t+1)} = \frac{ D_i^{(t)}\exp\{-\alpha_ty_ih_{t}(\textbf{x}_i)\}}{Z_t}$
		\State $[Output]: f_T{\textbf{(x)}}= sign  \left(\sum_{t=1}^T\alpha_th_t(\textbf{x})\right)$
	\end{algorithmic}
\end{algorithm}
\noindent Other boosting variations proposed are based on modifying the observation weighting function (\ref{eq:4}) (see, e.g., LogitBoost \citep{friedman2000additive}, MadaBoost \citep{domingo2000madaboost}, Gradient Boosting \citep{friedman2001greedy}, Stochastic Gradient Boosting \citep{friedman2002stochastic}, Local Boosting \citep{zhang2008local}), and although they address some of AdaBoost's limitations, they have not been proven to be strong PAC learners as formulated in Definition 1. The applications of boosting methods can be extended to regression, and multiclass problems easily, however as mentioned before we will focus only on the binary classification problem.  

\subsection{Random Forests}
\citet{breiman2001random} defines a random forest (RF) as an algorithm ``consisting of a collection of tree structured classifiers ${h_t(\textbf{x},\theta_t),t=1,...,T}$, where ${\theta_t}$ are independently and identically distributed random vectors." Each tree casts a unit vote for the most popular class at input $\textbf{x}$. 

RFs inject randomness by growing each of the $T$ trees on a random subsample of the training data, and also by using a small random subset of the predictors at each decision node split. The RF method is similar to boosting in the fact that it combines classifiers that have been trained on a subset sample or a weighted subset, but they differ in the fact that boosting gives different weight to the base learners based on their accuracy, while random forests have uniform weights. There has been ample research on these ensemble methods and how they perform under different settings. For a more complete review on their performance, the reader is referred to \cite{quinlan1996bagging, maclin1997empirical, dietterich2000ensemble}, and \cite{maclin2011popular}.

\subsection{Generalization Error Bound Based on Margins}
\citet{schapire98} proved an upper bound on the generalization error of an ensemble that does not depend on the number of classifiers combined $T$. The generalization error bound is formalized in Theorem 1. \\

\noindent
{\bf Theorem 1 }{\citep{schapire98}. \textit{Assuming that the base-classifier space $\mathscr{H}$ is finite, that is $\mathscr{\left|H\right|} < \infty$, and for any  $\delta> 0$ and $\theta>0$, then with probability at least $1-\delta$ over the training set $S$ with size $n$, every voting classifier $f$ satisfies the following bound:}}

\begin{equation}
\label{eq:5}
P\left[f_T(\textbf{x})\neq y\right] \leq \hat{P}\left[m(\textbf{x},y) \leq \theta) \right] + O\left( \frac{1}{\sqrt{n}}\sqrt{\frac{\ln{n}\ln{\mathscr{\left|H\right|}}} {\theta^2}+\ln{\frac{1}{\delta}}}{}\right),
\end{equation}

\noindent where $P\left[f_T(\textbf{x})\neq y\right]$ is the generalization error of the combined classifier, the term $\hat{P}\left[m(\textbf{x},y) \leq \theta\right]$ is the proportion of training set less than a value $\theta > 0$. When the hypothesis space is infinite, the expression $\ln{n}\ln{\mathscr{\left|H\right|}}$ is replaced by $d \log_2 (d/n)$, where $d$ is the VC-dimension of the space of all possible weak classifiers (a measure of complexity).  \citet{schapire98} use this bound to provide an explanation for the superior performance of AdaBoost, which they show is highly effective at increasing the margins. Based on this bound,  \citet{schapire98} concluded that larger margins should lead to lower generalization error, holding other factors constant, such as the cardinality of the hypothesis space $\mathscr{\left|H\right|}$, the sample size $n$, and $\delta$. This is sometimes referred to as the ``large margins theory" \citep{grove98, schapire98,  mason2000improved, shen2010boosting, wang2011refined, wang2012further,gao2013doubt, cid2012three, martinez2014role, zhou2014large}.  Given that maximizing the minimum margin has not yielded positive results in terms of generalization performance (see, e.g.,  \citealt{schapire98}, \citealt{grove98}), many authors have proposed optimizing other functions of the ensemble margin distribution instead. For instance, \cite{reyzin2006boosting} suggested maximizing the average or the median margin, while other researchers have proposed that minimizing the variance of the margins might be a key component in designing better performing ensembles (e.g., \citealt{shen2010boosting}).

\begin{figure} 
	\subfloat{\includegraphics[width=8.3cm, height=6.3cm]{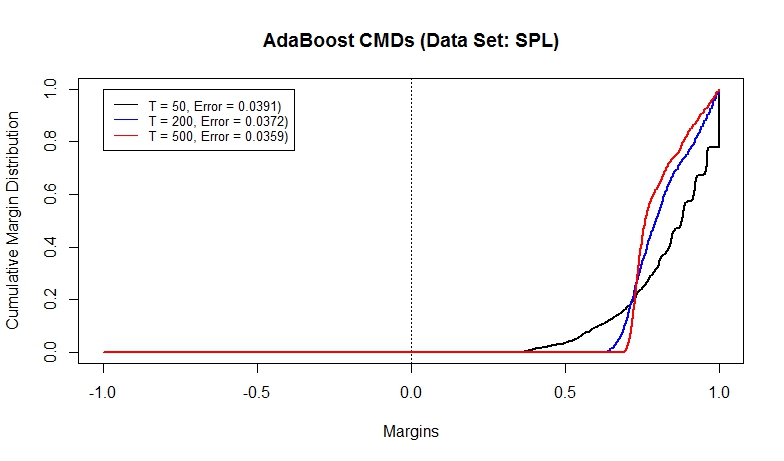}}\
	\subfloat{\includegraphics[width=8.3cm, height=6.3cm]{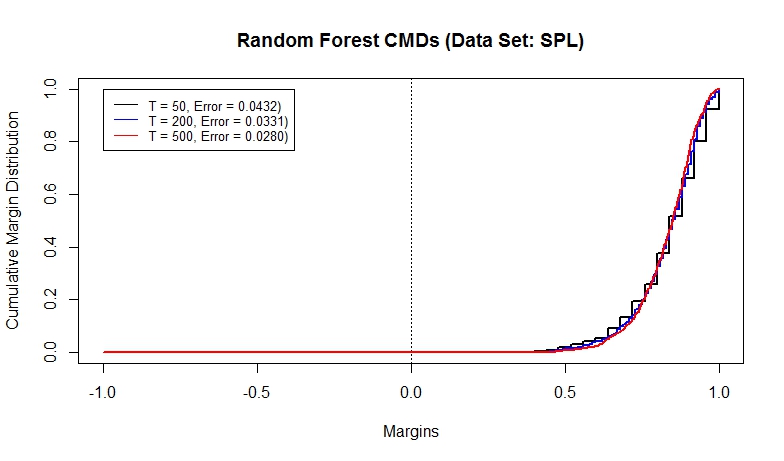}}\
	\caption{Cumulative Margin Distributions (CMDs) for AdaBoost and Random Forests for $T = 50, 200, 500$ for the SPL data set using full grown trees.}
	\label{fig:1}
\end{figure}

Figure \ref{fig:1} shows a typical behavior of the margins for a given ensemble. The cumulative margin distributions (CMDs) for ensembles of size $T = 50, 200, 500$ are shown for AdaBoost and random forests for the SPL data set (see table \ref{tab:2} for data set description) using full-grown trees. As the ensemble size grows, the test set error (which is an estimate of the generalization performance) rate decreases for both type of ensembles using these settings. More importantly, we can see that the variation of the margins does appear to decrease as $T$ increases. An important observation of Figure \ref{fig:1} is that, as \cite{schapire98} noted, ``boosting is especially aggressive at increasing the margins of the examples, so much so that it is willing to suffer significant reductions in the margins of those examples that already have large margins." This behavior is of particular significance, given that other researchers have shown that the lower margins play a pivotal role in ensemble performance (see, e.g., \citealt{guo2013margin}). The improvement in the lower margins is evident in Figure \ref{fig:1} as the ensemble size grows, and it also corresponds closely to better generalization performance. This is more markedly visible in AdaBoost than the random forest solution.  

\subsection{Relationship Between Ensemble Performance and Ensemble Diversity}
A different, but no less important measure of effectiveness, is how diverse the individual classifiers within an ensemble are.  Several researchers provide evidence on the importance of diversity within ensembles \citep{margineantu1997pruning, breiman2001random, kuncheva2003measures, liu2004empirical, banfield2005ensemble,  brown2005managing, blaser2016}. \citet{li2012diversity} proved an upper bound on the generalization error of any ensemble based on the diversity of its individual classifiers. The bound is given in Theorem 2. \\

\noindent
{\bf Theorem 2 }{\citep{li2012diversity}. \textit{Assuming that for every $f$, there exists a set of $T$ classifiers $h_t(\textbf{x}), t = 1,2,...,T$, satisfying $f_t(\textbf{x}) = 1/T \left(\sum_{t=1}^T h_t(\textbf{x})\right)$ and $div(f) \geq r$ for any i.i.d training set $S$ with size $n$, then for any $\epsilon$, and with probability at least $1-\delta$, for any $\theta > 0$, every function $f$ satisfies the following bound:}}

\begin{equation}
\label{eq:6}
P\left[f_T(\textbf{x})\neq y\right] \leq \hat{P}\left[m(\textbf{x},y) \leq \theta) \right] + \frac{C}{\sqrt{T}} \left( \frac{\ln{n} \ln{(T \sqrt{1/n +(1-1/n)(1-r)})}} {\theta^2} + \ln{\frac{1}{\delta}}\right),
\end{equation}

\noindent where $C$ is a constant, $div(f)$ is any measure of diversity or disagreement among voters of a given ensemble $f$. The bound in (\ref{eq:6}) cannot be minimized directly, but it suggests that an increase in diversity should improve generalization performance, holding other factors constant, such as the complexity of the classifier and sample size.  Note that this bound does depend on $T$.

\citet{margineantu1997pruning} define diversity or disimilarities of ensembles based on either the probability distributions on which the weak learner are derived, or the agreement (disagreement) of the classifiers in their predictions. Several researchers have used the $\kappa$  and $\kappa$-error diagrams proposed by \citet{margineantu1997pruning} to construct more diverse ensembles and/or evaluate their performance (see, e.g., \citealt{banfield2003new, kuncheva2004using, tsymbal2005diversity, rodriguez2006rotation, zhang2009novel}). \citet{germain2015risk} studied the relationship of ensemble diversity and the risk of majority voters, along with the first and second moments of the ensemble margins. They bounded the risk of a classifier with the expected disagreement between the individual learners. \citet{germain2015risk} define $M(\textbf{x},y)$ as a random variable, that given an example $(\textbf{x},y)$ drawn according to $P_{XY}$, outputs the margin of the majority voter on that example, that is:

\begin{equation}
\label{eq:7}
M(\textbf{x},y) = E_{XY}yf(\textbf{x}).
\end{equation}

\noindent The generalization error, or risk of $f$, can then be defined in terms of the margins, as the probability that the majority voter is incorrect $R[f] = P\left[M(\textbf{x},y) \leq 0 \right]$.  An important characteristic of the random variable $M(\textbf{x},y)$ is its first moment defined as:
\begin{equation}
\label{eq:8}
\mu_M = E_{XY}M(\textbf{x},y).
\end{equation}

\noindent $\mu_M$ is estimated by $\bar{m} = \sum_{i=1}^n m_i(\textbf{x}_i,y_i) /n$, where $m_i(\textbf{x}_i,y_i)$ is the margin of the $i^{th}$ training observation. As previously stated, several authors have concluded that maximizing $\bar{m}$, or maximizing the whole margins distribution should improve the generalization performance, and that simply maximizing the minimum margin ($\min_i{m_i(\textbf{x}_i,y_i)}$) does not result in improved generalization performance  \citep{schapire1999theoretical, reyzin2006boosting, grove98, zhou2014large}. The second moment of the distribution of $M(\textbf{x},y)$ is also of particular importance. We define the second moment as:
\begin{equation}
\label{eq:9}
\mu^2_M = E_{XY}M^2(\textbf{x},y).
\end{equation}

\noindent \citet{germain2015risk} provide an upper bound on the generalization error $R[f]$ of an ensemble that relates the diversity or expected disagreement between voters $(d_S)$, which is a particular measure of $\text{div}(f)$ on distribution $S$, and the second moment of the margin distribution $\mu^2_M$. The bound is given in Theorem 3. \\

\noindent
{\bf Theorem 3 }{\citep{germain2015risk}. For any distribution $Q$ on a set of voters and any distribution $D$ ox $X$, if $\mu_M>0$, we have:}

\begin{equation}
\label{eq:10}
P\left[f_T(\textbf{x})\neq y\right] \leq 1 - \frac{1-2R_D(G_Q)}{1-2d^D_Q},
\end{equation}

\noindent where $R_D(G_Q)$ is the Gibbs risk and $1-2d^D_Q$ relates the risk of the classifier with the second moment of the margin distribution. The reader is referred to \citet{germain2015risk} for a more complete explanation and the derivation of the bound, but the work in  \citet{germain2015risk} suggests that reducing the second moment $\mu^2_M$ of the margins of any given ensemble, should produce a more diverse and better performing classifier. Hypotheses presented by \cite{reyzin2006boosting}, \cite{shen2010boosting} and \citet{germain2015risk} all suggest that reducing the variation of the margins might also improve the generalization performance of an ensemble.  \cite{shen2010boosting},  for instance, proposed an algorithm named MD-Boost (Margin Distribution Boosting) that maximizes the average margin while reducing the variance of the margin distribution.  

\section{Existing Ensemble Pruning Methods}
\label{proposed}
The idea on diversity-based pruning is to reduce the size of a given ensemble based on the similarity of the weak learners, working on the premise that a more diverse ensemble performs better, however there are various types of ensemble pruning methods, not necessarily based on diversity. Most of them fall into either selection-based methods, or weight-adjusting methods \citep{chen2009predictive}. 

\subsection{Selection-Based Methods}
The main purpose of selection-based pruning methods is to either reject or select the given weak learner based on some criterion or criteria. The most common methodology used in selection-based pruning methods is to rank the weak learners within an ensemble according to some performance metric in a validation set and select a subset of the top $T_r$ out of the original $T$ weak learners. \citet{margineantu1997pruning} proposed several measures of diversity and ways to prune ensembles accordingly. They proposed the use of the Kullback-Leibler divergence (KL  distance) \citep{cover1991information} to prune ensembles, by maximizing the KL distance of the distribution upon which the classifiers were constructed. The KL distance between two probability distributions $p$ and $q$ is defined as:

\begin{equation}
\label{eq:11}
D(q\left|\right|p)\leq q \ln \frac{q}{p}+(1-q)  \ln \frac{1-q}{1-p}, 0\leq p,q\leq 1. 
\end{equation}     

\noindent To measure the agreement (or disagreement) of ensemble predictions, \citet{margineantu1997pruning}  use the Kappa statistic ($\kappa$) \citep{kohen1960coefficient}. Given two classifiers $h_a$ and $h_b$, \citet{margineantu1997pruning} consider their agreement by constructing a contingency table with elements $C_{ij}$, where $i \in \{-1,+1\}$ and $j \in \{-1,+1\}$, corresponding to the number of observations for which $h_a(\textbf{x}) = i$ and $h_b(\textbf{x}) = j$, and 

\begin{equation}
\label{eq:12}
\theta_1 = \frac{\sum_{i \in \{-1,+1\}}  C_{ii}}{n}.
\end{equation}

\noindent However, to account for class imbalances, the probability that the two classifiers agree is defined as:

\begin{equation}
\label{eq:13}
\theta_2= \sum_{i \in \{-1,+1\}}  \left(  \frac{\sum_{i \in \{-1,+1\}} C_{ij}}{n} \frac{\sum_{i \in \{-1,+1\}}  C_{ji}}{n}  \right).
\end{equation}

\noindent Finally, a measure of agreement $\kappa$ can be obtained by quantifying the likelihood to agree, compared to the expected agreement by chance:

\begin{equation}
\label{eq:14}
\kappa = \frac{\theta_1 - \theta_2}{1 - \theta_2}.
\end{equation}

\noindent The $\kappa$ measure in (\ref{eq:9}) has become a standard way to measure diversity in ensembles, and has been used extensively in selection-based pruning methods. Examples of other metrics used in selection-based methods include the margins of the ensemble and test set performance. For instance \cite{martinez2006pruning} use classification performance on a test set based on orientation ordering to select the best subensemble, while \cite{lu2010ensemble} and \cite{li2012diversity} proposed a heuristic to order the weak learners based on both their accuracy and their diversity. \cite{ prodromidis2001cost} proposed reducing the size of ensembles by minimizing a cost complexity metric. Ordering the weak learners could also be based on their margins (e.g. \citealt{guo2013margin}), or other measures that apply to specific types of analyses, such as time series (e.g., \citealt{ma2015several}). Other selection-based methods include formulating the selection of the pruned subsensemble as an integer optimization heuristic (e.g., \citealt{zhang2006ensemble}). The list of selection-based pruning methods presented here is not exhaustive, the user is referred to \cite{tsoumakas2009ensemble} for a more complete reference. One of the main limitations of most selection-based methods is that we must prespecify the size $T_r$ of the pruned subensemble. The selection-based approach is straightfoward, but does not guarantee the best possible subensemble and does not necessarily guarantee optimal performance. 																																																									

\subsection{Weight-Adjusting Methods}
For weight-adjusting pruning methods, the main goal is not necessarily to prune the ensemble, but to adjust the weights of the weak learners, so that the generalization error is improved. In the process, some of the weights get zeroed out and consequently the ensemble is reduced (pruned). The main limitation of most weight-adjusting methods is that they do not have any theoretical guarantee to diminish the size of the ensemble, nor is there any explicit formulation to do so in their heuristics. For example \cite{grove98} used a linear programming technique to adjust the weights of the weak learners, so that the ensemble's minimum margin is maximized.  \cite{grove98} remark that the final ensemble was generally reduced significantly with their proposed algorithm. \cite{demiriz2002linear} also used a weight-adjusting linear program to optimize a generalization error bound, which they show is also effective at pruning the ensemble. \cite{chen2009predictive} use expectation propagation to approximate the posterior estimation of the weight vectors. There are other weight-adjusting pruning methods in the literature that merit attention (e.g., \citealt{chen2006probabilistic}), and the reader is again referred to \cite{tsoumakas2009ensemble} for a more exhaustive reference.

\section{QMM (Quadratic Margin Maximization) Optimization Algorithm}
\label{proposed}
In this section, we propose a weight-adjusting method based on quadratic programming to reduce the fraction of weak learners utilized in a particular ensemble. In designing our proposed algorithm, we take into account the generalization error bound in Theorem 1, which suggests that larger margins should lead to lower generalization error. Specifically, we focus on increasing the lower margin percentiles \citep{schapire98}. We also consider Theorem 2, which relates the performance of ensembles to the diversity of classifiers, with higher diverse classifiers expected to perform better, along with Theorem 3, which states that reducing the variance of the margins induces a more diverse and better performing combined classifier. The quadratic program formulation aims to tune the weights of the given ensemble, such that the pairwise correlations of the weak learners  and the variance of the margins are minimized, while maximizing the lower percentiles of the margins. 

\subsection{Quadratic Programming (QP) Formulation}
We assume that we are given an ensemble solution, i.e., a set of weak learners  $h_t(\textbf{x}), t = 1,2,...,T$, and a set of weights $\boldsymbol\alpha \in (0,1)^T$, where $\boldsymbol\alpha=\{\alpha_1   \alpha_2  \dots  \alpha_T\}^\top$, where $0\leq \alpha_t \leq 1$ and $\sum_{t=1}^T \alpha_t = 1$, associated with the weak classifiers. The weights $\alpha_t$ can be normalized without loss of generality. Note that for the training sample $S=\{( \textbf{x}_i, y_i ), i = 1,2, ...,n \}$ used to produce the ensemble solution, the values of the weak learner predictions $h_{it}$ and the weights $\alpha_t$ are fixed. We will let \(h_{it}=\pm1\) denote the prediction of the $t^{th}$ weak learner for the $i^{th}$ observation in the training data, $\textbf{y} = \{y_1 y_2 \dots y_n\}^\top$ and $\textbf{m} =\{m_1 m_2 \dots m_n\}^\top$ . We define the matrix 
\begin{equation}
\label{eq:15}
\textbf{H}_{n,T} = \begin{pmatrix}
h_{11} & h_{12} & \cdots & h_{1T} \\
h_{21} & h_{22} & \cdots & h_{2T} \\
\vdots  & \vdots  & \ddots & \vdots  \\
h_{n1} & h_{n2} & \cdots & h_{nT}
\end{pmatrix},
\end{equation}
\noindent where $\textbf{H} \in \{-1,1\}^{n \times T}$, as the predictions matrix for the $T$ weak classifiers within the ensemble. The error matrix $\textbf{E}=\textbf{H} \bullet \textbf{y}$ is defined as:

\begin{equation}
\textbf{E}_{n,T} = \begin{pmatrix}
y_1h_{11} & y_1h_{12} & \cdots & y_1h_{1T} \\
y_2h_{21} & y_2h_{22} & \cdots & y_2h_{2T} \\
\vdots  & \vdots  & \ddots & \vdots  \\
y_nh_{n1} & y_nh_{n2} & \cdots & y_nh_{nT}
\end{pmatrix},
\end{equation}

\noindent where $\textbf{E} \in \{-1,1\}^{n \times T}$, with element $E_{ij} = -1$ when the prediction of $j^{th}$ classifier for observation $i$ is incorrect, $+1$ otherwise. Note that $\textbf{E}\boldsymbol\alpha = \textbf{m}$.  Let $\hat{\boldsymbol\Sigma} = \text{cov}(\textbf{E})$ be the sample covariance matrix of $\textbf{E}$, where $\hat{\boldsymbol\Sigma}$ is a symmetric positive definite matrix, with error variance of the each weak learner $h_t$ in the diagonals $\hat{\sigma}_{h_{t}}^2$, and the errors covariance for weak learners $i$ and $j$ in the off-diagonals  $\hat{\sigma}_{h_{ij}}$.

\begin{equation}
\hat{\boldsymbol\Sigma}_{T,T} = \begin{pmatrix}
\hat{\sigma}_{h_{1}}^2 & \hat{\sigma}_{h_{12}}  & \cdots & \hat{\sigma}_{h_{1T}} \\
\hat{\sigma}_{h_{21}} & \hat{\sigma}_{h_{2}}^2 & \cdots & \hat{\sigma}_{h_{2T}} \\
\vdots  & \vdots  & \ddots & \vdots  \\
\hat{\sigma}_{h_{T1	}} & \hat{\sigma}_{h_{T2}} & \cdots & \hat{\sigma}_{h_{T}}^2
\end{pmatrix}.
\end{equation}

\noindent Let $m_{(1)} < m_{(2)} < \cdots < m_{(n)}$ denote the ensemble margins after arrangement in increasing order of magnitude, and the margin vector $\boldsymbol\varphi_{\upsilon} \in (-1,1)^{\ceil{n \upsilon}} = \{m_{(1)} m_{(2)} \cdots m_{(\ceil{n \upsilon})}\}^\top$ for a given percentile $\upsilon$.  We define the matrix $\Lambda_{\upsilon} \in (-1,1)^{\ceil{n \upsilon} \times T}$ as

\begin{equation}
\boldsymbol\Lambda^{\upsilon}_{\ceil{n \upsilon},T} = 
\begin{pmatrix}
y_{(1)}h_{(11)} & y_{(1)}h_{(12)} & \cdots & y_{(1)}h_{(1T)} \\
y_{(2)}h_{(21)} & y_{(2)}h_{(22)} & \cdots & y_{(2)}h_{(2T)} \\
\vdots  & \vdots  & \ddots & \vdots  \\
y_{(\ceil{n \upsilon})}h_{(\ceil{n \upsilon}1)} & y_{(\ceil{n \upsilon})}h_{(\ceil{n \upsilon}2)} & \cdots & y_{(\ceil{n \upsilon})}h_{(\ceil{n \upsilon}T)}
\end{pmatrix},
\end{equation}
such that
\begin{equation*}
\boldsymbol\Lambda^{\upsilon} \begin{pmatrix}
\alpha_1 \\ \alpha_2 \\ \vdots \\  \alpha_T
\end{pmatrix} 
= \begin{pmatrix}
m_{(1)} \\ m_{(2)} \\ \vdots \\ m_{(\ceil{n \upsilon})}
\end{pmatrix}.
\end{equation*}
The main goal is to minimize the error covariance matrix $\hat{\boldsymbol\Sigma}$, such that the lower $\upsilon$ margin percentiles are optimized.  We call the quadratic program used to achieve this solution the QMM algorithm and it can be expressed in the following form: 

\begin{equation}
\begin{aligned}
& {\text{minimize}}
& & \textbf{w}^\top \hat{\Sigma} \textbf{w}\\
& \text{subject to}
& & \boldsymbol\Lambda^{\upsilon}\textbf{w} \geq \boldsymbol\varphi^{\upsilon},
& & \textbf{w}^\top \textbf{1} = 1,
& & \textbf{w}\geq \textbf{0},
\end{aligned}
\end{equation}
where $\textbf{w}$ are the new weights for the weak learners to be determined by solving the QP.  The constraint $\Lambda^{\upsilon} \textbf{w} \geq \boldsymbol\varphi^{\upsilon}$ guarantees that the choice of $\textbf{w}$ generates lower $\upsilon$ margin percentiles $\boldsymbol\Lambda^{\upsilon} \begin{pmatrix} w_1 & w_2 & \cdots &  w_T \end{pmatrix}^\top  $ at least as large, or larger than the lower $\upsilon$ margin percentiles generated by the ensemble $\boldsymbol\Lambda^{\upsilon} \begin{pmatrix} \alpha_1 & \alpha_2 & \cdots &  \alpha_T \end{pmatrix}^\top$. Optimal solutions generally induce some of the $\textbf{w}$ to equal zero, therefore reducing the final ensemble size, but the zeroing-out of the weights $\textbf{w}$ is not guaranteed, as in most weight-adjusting pruning methods. The main hypothesis here is that reducing the error covariance, and minimizing the margin variance will cause the optimization algorithm to zero-out weights corresponding to the less diverse, worse performing weak learners. 

One of the main issues that we can run into with QMM algorithm is the possibility of a less than full-rank $\textbf{H}$ matrix because of column dependencies, which consequently would render the $\hat{\boldsymbol\Sigma}$ matrix also rank-deficient, and not positive semidefinite.  \cite{chen2009predictive} use the least square pruning method ($lscov$) as a baseline before their proposed method to alleviate the rank deficient cases. We propose using a QR decomposition with pivoting to detect the column dependencies of $\textbf{H}$ . If $\textbf{H}$  has rank $r < \min{(n,T)}$, then there is an orthogonal matrix $\textbf{Q}$ and a permutation matrix $\textbf{P}$ such that
  
\begin{equation*}
\textbf{Q}^\top \textbf{H} \textbf{P} = 
\begin{pmatrix}
\textbf{I}_{r} & \textbf{0}_{r,T-r}  \\
\textbf{0}_{n-r, r} & \textbf{0}_{n-r,T-r} & \\
\end{pmatrix},
\end{equation*}
The $r$ column pivot positions for $\textbf{H}$ with indices $D_H = \{d_1, d_2, \cdots,d_r\}$ determine a basis for the number of columns needed to span the ensemble space of $T$ classifiers, while the $T-r$ non-pivot columns with indices $G = \{g_1, g_2, \cdots, g_{T-r}\}$ can be expressed as linear combinations of the $r$ pivot columns. Let $\boldsymbol\alpha'=\{\alpha'_{1}   \alpha'_{2}  ...  \alpha'_{r}\}^\top= \{\alpha_{d_1}   \alpha_{d_2}  ...  \alpha_{d_r}\}^\top$ denote the weights of the classifiers for the $r$ column pivot positions of $\textbf{H}$. Dropping the remaining $T-r$ columns of $\textbf{H}$, we obtain:
\begin{equation}
\textbf{H}'_{n,r} = \begin{pmatrix}
h'_{11} & h'_{12} & \cdots & h'_{1r} \\
h'_{21} & h'_{22} & \cdots & h'_{2r} \\
\vdots  & \vdots  & \ddots & \vdots  \\
h'_{n1} & h'_{n2} & \cdots & h'_{nr}
\end{pmatrix} = \begin{pmatrix}
h_{1d_1} & h_{1d_2} & \cdots & h_{1d_r} \\
h_{2d_1} & h_{2d_2} & \cdots & h_{2d_r} \\
\vdots  & \vdots  & \ddots & \vdots  \\
h_{nd_1} & h_{nd_2} & \cdots & h_{nd_r}
\end{pmatrix},
\end{equation}
where $\textbf{H}'$ is a full-rank matrix comprised $r \leq T$ classifiers, therefore also potentially pruning the ensemble. We also define the matrix  $\textbf{H}^* = \textbf{H}/\textbf{H}'$ as
\begin{equation}
\textbf{H}^*_{n,T-r} = \begin{pmatrix}
h^*_{11} & h^*_{12} & \cdots & h^*_{1T-r} \\
h^*_{21} & h^*_{22} & \cdots & h^*_{2T-r} \\
\vdots  & \vdots  & \ddots & \vdots  \\
h^*_{n1} & h^*_{n2} & \cdots & h^*_{nT-r}
\end{pmatrix} = \begin{pmatrix}
h_{1g_1} & h_{1g_2} & \cdots & h_{1g_{T-r}} \\
h_{2g_1} & h_{2g_2} & \cdots & h_{2g_{T-r}} \\
\vdots  & \vdots  & \ddots & \vdots  \\
h_{ng_1} & h_{ng_2} & \cdots & h_{ng_{T-r}}
\end{pmatrix}.
\end{equation}
Let $\textbf{h}'_j, j = 1,\cdots,r$ denote the $j^{th}$ column of $\textbf{H}'$ with $\alpha'_r$ associated weights, and $\textbf{h}^*_k, k = 1,\cdots,T-r$ the $k^{th}$ column of $\textbf{H}^*$ with $\alpha^*_{T-r}$ associated weights. If $r < \min{(n,T)}$ then $\textbf{h}'_j = \sum^{G_R}_{i = 1}(c_i \textbf{h}'_g)$, where $G_R$ is the number of columns from $\textbf{H}^*$ equal to $\textbf{h}'_j$. Finally, the $j^{th}$ element of the weight vector for the classifier $\boldsymbol\alpha'$ would be defined as:
\begin{equation}
\label{eq:22}
\alpha'_j = \sum^{G_R}_{i=1} (\alpha^*_i).
\end{equation}
We would use $\textbf{H}'$ and $\boldsymbol\alpha'$ instead of $\textbf{H}$ and $\boldsymbol\alpha$ in the QMM  optimization formulation in (19) as a baseline, and the covariance matrix $\hat{\boldsymbol\Sigma}' = \text{cov} (\textbf{H}' \bullet \textbf{y})$ with the corresponding $\boldsymbol\Lambda'^{\upsilon}$ and $\boldsymbol\varphi'^{\upsilon}$. Therefore even if $\textbf{H}$ is less than full-rank, the optimization would still be based on a full-rank $\hat{\boldsymbol\Sigma}'$ matrix. It is also possible that $\hat{\boldsymbol\Sigma}$ is a full rank matrix but not positive semidefinite, especially in the case where $n<T$. In that case, the QMM algorithm will fail to find an optimal solution, and the resulting ensemble solution will be the original $f_T(\textbf{x})= \text{sign}(\textbf{E} \boldsymbol\alpha)$ or most likely $f_T(\textbf{x})= \text{sign}(\textbf{E}' \boldsymbol\alpha')$.  All the steps of the QMM algorithm are summarized in Algorithm \ref{alg2}.

\begin{algorithm}
	\caption{QMM Algorithm}\label{euclid}
	\label{alg2}
	\begin{algorithmic}[1]
		\State $[Input]:  S$, $\textbf{H} = \{h_{it}\}, t= 1,...,T$, $\boldsymbol\alpha=\{\alpha_1   \alpha_2  ...  \alpha_T\}^\top$, $\textbf{m}$, and $\upsilon$ from Algorithm 1
		\State $[Compute]: \textbf{H}'$, $\boldsymbol\alpha'$, $\hat{\boldsymbol\Sigma}$, and $\boldsymbol\Lambda^{\upsilon}$. Let $\textbf{H} \doteq \textbf{H}'$, $\boldsymbol\alpha \doteq \boldsymbol\alpha'$, $\boldsymbol\Lambda^{\upsilon} \doteq \boldsymbol\Lambda'^{\upsilon}$, and $\boldsymbol\varphi^{\upsilon} \doteq \boldsymbol\varphi'^{\upsilon}$
		\State $[Optimize]:$ minimize $\textbf{w}^\top \hat{\boldsymbol\Sigma} \textbf{w}$ subject to $\boldsymbol\Lambda^{\upsilon} \textbf{w} \geq \boldsymbol\varphi^{\upsilon}, \textbf{w}^\top \textbf{1} = 1, \textbf{w}\geq 0$
		\State $[Return]: \textbf{w}$
	\end{algorithmic}
\end{algorithm}

\noindent We illustrate the performance of the QMM algorithm in Figure \ref{fig:2} by plotting the cumulative margin distributions (CMDs) for the original AdaBoost and random forest solutions and the proposed QMM method on the Breast Cancer (BC) data set (see table 1 for data set description). The QMM ensemble uses only 74 trees (4-node, depth = 2) out of the 200 used by the AdaBoost solution, and 76 trees out of the random forest solution of 200 trees with equal error rates.

\begin{figure} 
	\subfloat{\includegraphics[width=8.3cm, height=6.3cm]{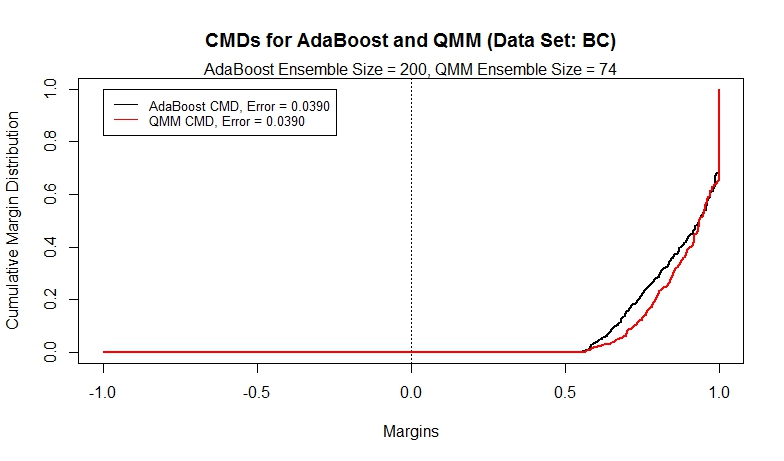}}\
	\subfloat{\includegraphics[width=8.3cm, height=6.3cm]{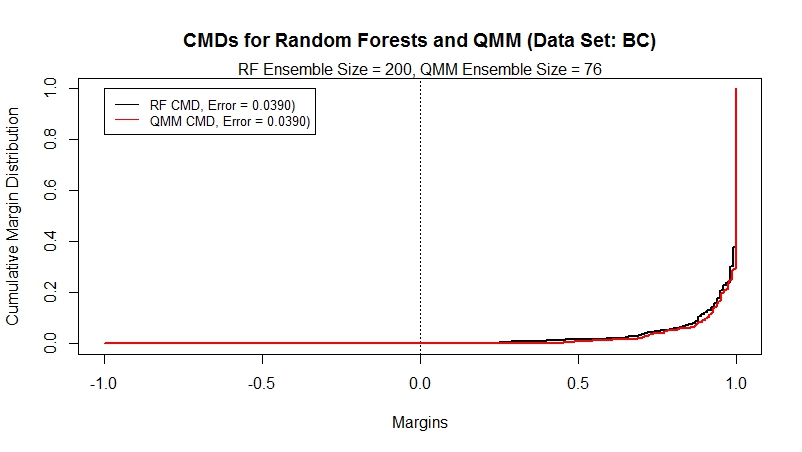}}\
	\caption{By round comparison of AdaBoost and MMI algorithm for the Splice data set using decision trees with 4 terminal nodes}
	\label{fig:2}
\end{figure}

\subsection{Choice of $\upsilon$}
The parameter $\upsilon$ determines the fraction of margins that will be maximized, that is, the QMM solution will constraint these margins to be at least as large as those of the original ensemble solution. Selecting a high value of $\upsilon$ will require a higher fraction of margins to be improved or maintained and since AdaBoost and random forests are highly effective at increasing the training margins \citep{schapire98}, the optimization will likely fail to find a feasible set. For $\upsilon = 1$ we would likely obtain the original solution $\textbf{w} = \boldsymbol\alpha$ or an even an infeasible solution in the optimization algorithm, in which case we also assign $\textbf{w} = \boldsymbol\alpha$. On the other hand, setting $\upsilon$ close to 0 would most likely result in a smaller but underperforming ensemble. The selection of $\upsilon$ can be based on cross-validation, but this will result in higher computational costs. Making use of the structure of the margins distribution for the particular ensemble might also give some useful insights. For instance, \cite{wang2011refined} developed an upper bound based on a single margin instance called the equilibrium margin (EMargin) as an explanation of the performance of ensemble methods. To explain the EMargin, we use the Bernoulli Kullback-Leiler function $D(q\left|\right|p)$ as defined in (\ref{eq:6}). $D(q\left|\right|p)$ is a monotone increasing function for a fixed $q$ and $q \leq p < 1$. We can also see that $D(q\left|\right|p)=0$  when $p = q$ and $D(q\left|\right|p) \rightarrow \infty$  as $p\rightarrow 1$. The bound that relates the EMargin to the performance of an ensemble classifier is presented in Theorem 4.
\\

\noindent
{\bf Theorem 4} {\it \citep{wang2011refined}. Assuming that the base-classifier space $\mathscr{\left|H\right|}$ is finite, and for any  $\delta> 0$ and $\theta >0$, then with probability at least $1-\delta$ over the training set $S$ with size $n$, every voting classifier $f$ satisfies the following bound:}
\\

\begin{equation}
\label{eq:9}
P\left[f_T(\textbf{x})\neq y\right] \leq \frac{\ln{\mathscr{\left|H\right|}}}{n}+\inf_{q\in \left(0, \frac{1}{n}, \frac{2}{n},...,1\right)}D^{-1} \left(q;u \left[\hat{\theta}(q)\right]\right),
\end{equation}

\noindent where $u\left[\hat{\theta}(q)\right]= \frac{1}{n} \left( \frac{8\ln\mathscr{\left|H\right|}}{\hat{\theta}^2(q)} \ln \left(\frac{2n^2}{\ln \mathscr{\left|H\right|}}\right)+\ln{\mathscr{\left|H\right|}}+\ln \frac{1}{\delta} \right)$ and 

\begin{equation}
\label{eq:10}
\hat{\theta}(q)= \sup_{\theta \in \left( \sqrt{\frac{8}{\mathscr{ \left| H \right|}}},1\right]}  \hat{P}\left[m(\textbf{x},y) \leq \theta) \right] \leq q.
\end{equation}
The optimal value of $q$ in (\ref{eq:9}) defined as $q^*$ and evaluated at $\hat \theta(q^*)$ is called the EMargin, while $q^*$  is called the EMargin error. \cite{wang2011refined} suggest that an ensemble with higher EMargin $\hat \theta(q^*)$ and lower EMargin error $q^*$ should perform better, holding everything else constant. With these results in mind, the value of $\upsilon$ can be set to $\hat{P}\left[m(\textbf{x},y) \leq \hat \theta(q^*) \right] = \upsilon$. 

The main drawback of using the EMargin to select the value of  $\upsilon$ is the extra computational cost, as well as the difficulties in obtaining the size of the base-classifier space $\mathscr{\left|H\right|}$. \cite{wang2011refined} suggested using a preespecified number of thresholds uniformly distributed on $\left[0,1\right]$ on each feature with a fixed classifier complexity, such as decision stumps, so that $\mathscr{\left|H\right|}$ can be computed accurately.   
\begin{figure} 
	\subfloat{\includegraphics[width=8.6cm, height=5.5cm]{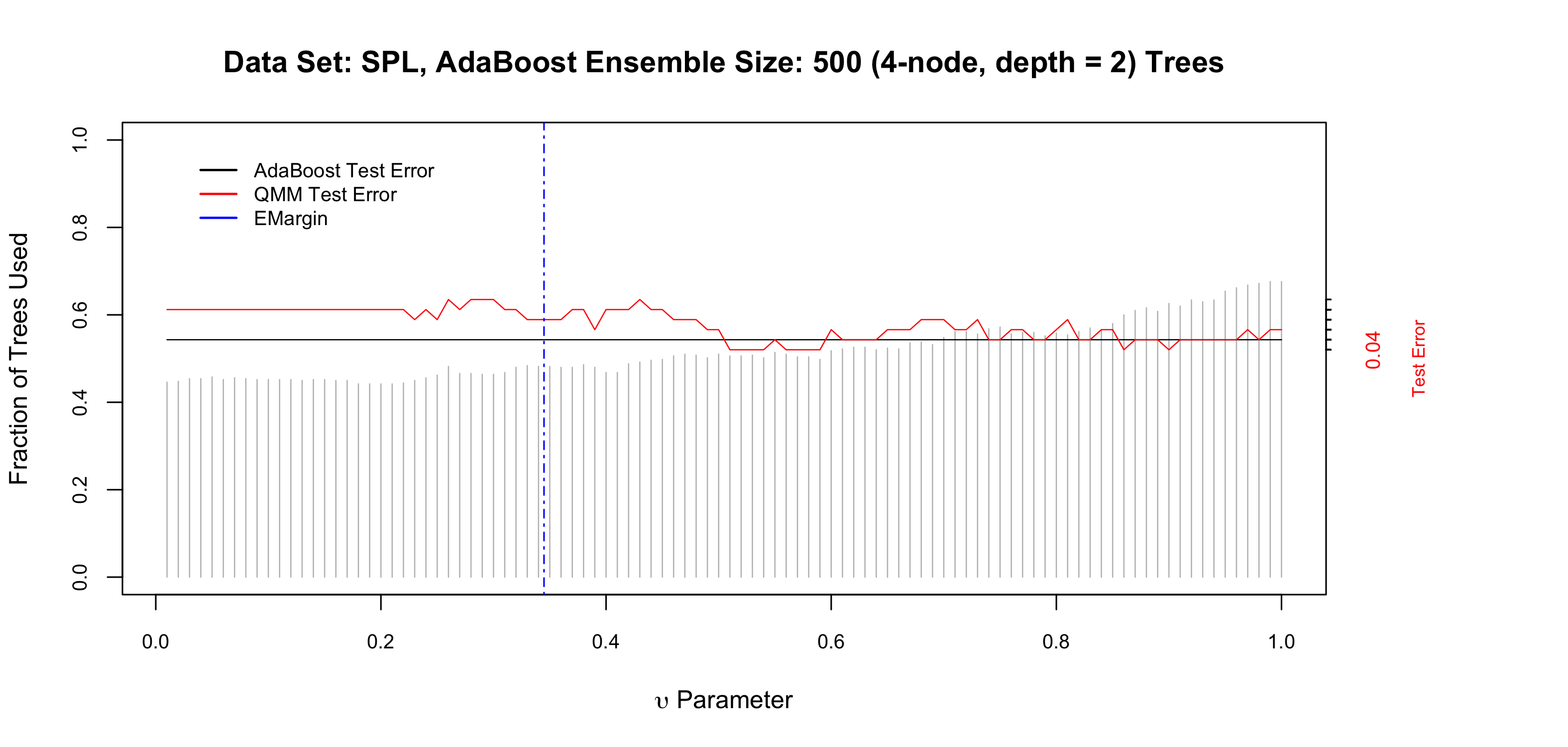}}\
	\subfloat{\includegraphics[width=8.6cm, height=5.5cm]{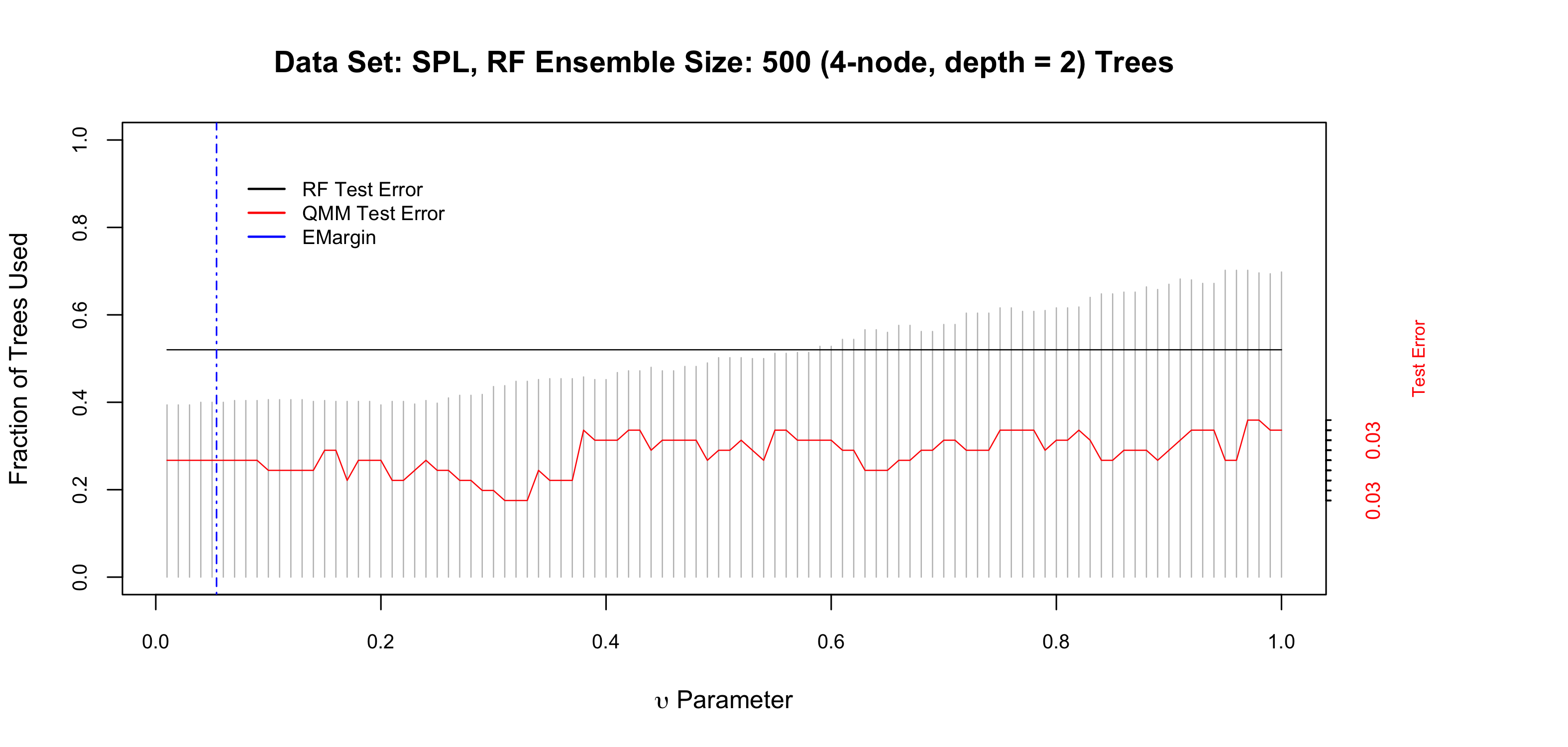}}\
	\caption{Performance of the QMM algorithm under different values of  $\upsilon$ for AdaBoost and Random Forests using full-grown trees. The shaded grey represents the fraction of weak learners used given the ensemble size, while the red and black curves represent the test set error rates of the QMM algorithm and AdaBoost respectively.}
	\label{fig:2}
\end{figure}
Figure 3 illustrates the performance of the QMM algorithm for the SPL data set under different values of  $\upsilon$ for both AdaBoost and random forests. We fix the complexity and obtain the size of hypothesis space by using 4-node (depth = 2) decision trees. We further normalize each feature of the SPL data set to $\left[0,1\right]$ and consider only 100 thresholds uniformly distributed on $\left[0,1\right]$ on each feature, so that  $\mathscr{\left|H\right|} = \left(2 \times 100 \times p\right)^3$, where $p =60$. We then obtain the fraction of trees used, the test set error rate for values of $\upsilon = (0,1]$ with increments of $0.01$. This specific example shows that lower values of  $\upsilon$ correspond to the highest ensemble pruning rate for both AdaBoost and random forests, however it translates to worse performance for the AdaBoost algorithm. For higher values of $\upsilon$, it becomes increasingly more difficult for the algorithm to prune the ensemble and consequently the QMM algorithm yields a higher fraction of trees selected. It is clear that the optimal value of $\upsilon$ depends on the ensemble type, as well as the data set used. The vertical dotted blue line in Figure 3 corresponds to the EMargin. The example in Figure 3 is a typical performance of the QMM algorithm and shows that the  performance of the QMM algorithm is better for higher values of $\upsilon$ with an AdaBoost solution, while values of $\upsilon$ ranging from 0.01 to 0.40 result in improved performance in random forests. Using the EMargin to set the parameter $\upsilon$ generally results in a higher pruning rate with acceptable performance results, however to reduce computational costs in our simulations, we have set $\upsilon = 0.5$ for AdaBoost, and  $\upsilon = 0.25$ for random forests knowing that this value could be further optimized by different means that not are limited to the use of the EMargin, but can also be derived by cross validation, especially if computational costs for the specific problem are not an issue. For further simulations, we do not restrict the decision trees to a preespecified number of thresholds.

\begin{table}
	\centering
	\caption{Description of Data Sets (Low Dimensional)}
	\begin{tabular}{lllccc}
		\toprule
		Data Set & Description &  Source & Train & Test  & Features  \\
		\midrule
		AU & Australian Credit Approval&\cite{Lichman2013} &482 &  208       & 14      \\
		BAN & Banana Data Set &\cite{ratsch} &3710&  1590       & 2      \\
		BC & Breast Cancer Wisconsin & \cite{Lichman2013}  &478   & 205      & 10    \\
		DIA & Diabetes Patient Records & \cite{Lichman2013} &537   & 231      & 8     \\
		FC & Fourclass Non-Separable  & \cite{ho1996building} &603   & 259      & 2     \\	
		\bottomrule
	\end{tabular}%
	\label{tab:1}%
\end{table}%

\section{Experiments and Simulations}
\label{proposed}
The application of the QMM algorithm is illustrated on several synthetic and real data sets to gauge its applicability. We compare the performance of QMM algorithm to the original ensemble using the test set error rate and the percentage of the weak learners used at a given ensemble size. Tables \ref{tab:1}, \ref{tab:2} and \ref{tab:3} show descriptions and properties of the data sets utilized in the simulations. Data sets that do not contain a validation data set were split on a 70/30 sampling scheme. The data sets range in size and dimensionality. We have arbitrarily broken down the data sets into three main types: low dimensional data sets (Table \ref{tab:1}), mid dimensional data sets (Table \ref{tab:2}), and high dimensional data sets (Table \ref{tab:3}). The ensembles used for the this analysis are AdaBoost and random forests. If the QMM algorithm does not result in a feasible solution we have provided alternative decreasing values of $\upsilon$. For AdaBoost, we have set the following options $\upsilon \in \{0.50, 0.25, 0.05, 0.01\}$, and for random forests $\upsilon \in \{0.25, 0.15, 0.05, 0.01\}$.
\subsection{QMM Performance}
\label{proposed}

\begin{table}
	\centering
	\caption{Description of Data Sets (Mid Dimensional)}
	\begin{tabular}{lllccc}
		\toprule
		Data Set & Description &  Source & Train & Test  & Features  \\
		\midrule
		IJC & IJCNN 2001 Competition & \cite{prokhorov2001slide}  &49990 & 91701 & 22   \\
		ION & Ionosphere Data Set & \cite{Lichman2013}  &245 &106  & 34    \\
		MR & Mushrooms Data Set& \cite{Lichman2013}  &5686  & 2438   & 112   \\
		SON  & Sonar Data Set  & \cite{Lichman2013} &145  & 63  & 60    \\
		SPL  & DNA Splice Junctions   & \cite{Lichman2013} &1000  & 2175  & 60   \\		
		\bottomrule
	\end{tabular}%
	\label{tab:2}%
\end{table}%

Figure \ref{fig:3}  illustrates the test error curves and percentage of trees utilized for the QMM algorithm versus the original AdaBoost on low dimensional data sets. The ensemble is grown to a size $T = 500$. The shaded grey area represents the fraction of weak learners used for the given ensemble size, while the red and black curves represent the test set error rates of the QMM algorithm and AdaBoost respectively. We have used different topologies of CART classification trees \citep{breiman1984classification} that include decision stumps, 4-node (depth = 2) trees and full-grown trees as our base-learning classifiers. We can see in Figure \ref{fig:3} that the test set error rates of the QMM algorithm are comparable and sometimes outperform those of AdaBoost. The QMM algorithm only uses a fraction of the classifiers given by the AdaBoost solutions. As $T$ increases, the number of decision trees selected by the QMM algorithm decreases. This happens more markedly for ensembles of decision stumps. For instance, for low dimensional data sets on average only 22 trees out of an ensemble of size $T=500$ are assigned positive weights by the QMM algorithm, a 95.6\% reduction. As the complexity of the classifier increases, a higher diversity is expected within the ensemble members. For 4-node trees, the ensemble is on average pruned to only 40.2 out of 500 trees, which suggests that the fraction of trees used by the QMM algorithm is generally higher as the tree depth increases. This is likely due to the fact that a less complex tree topology will induce more similar resulting trees. For full-grown trees the QMM algorithm prunes the AdaBoost ensemble to an average of 120 trees out of 500.  The performance of the QMM algorithm in mid dimensional data sets does not differ significantly from low dimensional data sets as illustrated in Figure \ref{fig:4} . On average 37.8, 134 and 149.6 trees out 500 were selected for decision stumps, 4-node trees and full-grown trees respectively, which might also suggest lower pruning rates as the dimensionality of the data set increases. Results also indicate a similar performance of the resulting ensemble of the QMM algorithm compared to AdaBoost in terms of generalization ability.  An interesting phenomenon happens for ION and SON data sets, showing a sudden increase in the percentage of trees used right after $T > n$, and this is likely due to the fact that $\hat{\boldsymbol\Sigma}' $ is not positive semidefinite for data sets with $n < T$, nevertheless the QR decomposition of the $\textbf{H}$ matrix reduces the size of the ensemble even if the QP formulation fails to find any feasible solution to prune the ensemble further. Figure \ref{fig:5} shows the performance of the proposed method for high dimensional data sets, which suggests a similar story to the low and mid dimensional data sets simulations. On average 67.3, 164.7 and 327 trees out 500 were used for decision stumps, 4-node trees and full-grown trees respectively. The trend further suggests that as the dimensionality of the data set increases, a higher number of trees is needed to achieve feasibility in the constraints of the QMM algorithm. An interesting note for the simulations on AdaBoost ensembles is that on occasions the percentage of trees varies wildly as $T$ increases for some data sets and this is due to the algorithm going back and forth to different values of $\upsilon$.

Figures \ref{fig:6},  \ref{fig:7} and  \ref{fig:8} show how the QMM algorithm compares to random forest ensembles for low, mid and high dimensional data sets using ensembles of size $T = 1,2,...,500$. We can see the fraction of weak learners used in the QMM algorithm is also on average decreasing as $T$ increases, and the performance is similar to the results using AdaBoost ensembles.  For low dimensional data sets the QMM algorithm uses on average 17, 57.2 and 216 trees out of 500 for decision stumps, 4-node trees and full-grown trees respectively. The average tree sizes for the QMM algorithm are 31.2, 97.6 and 186.8 for decision stumps, 4-node trees and full-grown trees respectively for mid dimensional data sets, and 119.3, 202.7 and 273.7 for high dimensional data sets.   The QMM algorithm is able to produce significant improvements to the random forest solutions in some data sets, however it does perform worse in others. On average, though, the QMM algorithm does provide improvements both in the test set error and the reduction in the size of the ensemble compared to the random forest solutions.

\begin{table}
	\centering
	\caption{Description of Data Sets (High Dimensional)}
	\begin{tabular}{lllccc}
		\toprule
		Data Set & Description &  Source & Train & Test  & Features  \\
		\midrule
		CC & Colon Cancer Data & \cite{alon1999broad}  &43   &  19     & 2000    \\
		GIS & Digit Recognition Data &\cite{guyon2004result}  &6000  & 1000  & 5000   \\
		MAD & Artificial Data & \cite{guyon2004result}  &2000  & 600   & 500   \\	
		\bottomrule
	\end{tabular}%
	\label{tab:3}%
\end{table}%

\subsection{QMM Performance Under Noise}

Ensembles, particularly AdaBoost, tend to be sensitive to outliers and noise. \cite{grove98}, \cite{mason2000improved} and \cite{dietterich2000ensemble} provide evidence that AdaBoost does overfit and the generalization error deteriorates rapidly when the data is noisy. \cite{long2010random} proved that for any boosting algorithm with a potential convex loss function, and any nonzero random classification noise rate, there is a data set, which can be efficiently learnable by the booster if there is no noise, but cannot be learned with accuracy better than 1/2 with random classification noise present. Many methods that automatically handle noisy data and outliers have been proposed to alleviate the limitations of AdaBoost. Algorithms such as BrownBoost \citep{freund2001adaptive}, LogitBoost \citep{friedman2000additive}, MadaBoost \citep{domingo2000madaboost}, LPReg-AdaBoost \citep{ratsch2001soft}, $\nu$-LP and $\nu$-ARC, \citep{Ratsch01robustboosting} mostly attempt to accommodate noise by somehow allowing unusual observations to fall in on the wrong side of the prediction in subsequent iterations of AdaBoost. Although bagging and random forests generally perform better than AdaBoost under noisy circumstances, they are still not completely robust to noise and outliers \citep{maclin1997empirical}. Deleting outliers (also called noise filtering or noise peeling) by pre-processing the data is preferable under certain high noise circumstances \citep{martinez2016noise}. 		

To visually illustrate how the QMM algorithm performs under a noisy scenario, we generate a synthetic two dimensional data set consisting of 1000 data points uniformly distributed on the unit square with $y=-1$ when $x1+x2< 1$, and $y = +1$ when $x1+x2> 1$, and randomly assign to $y \in \{-1,1\}$, when $x1+x2=1$.  The true boundary would be the diagonal line $x1+x2 = 1$. To generate noise, we randomfly flip the response variable of 20 observations with $y = -1$ to $y = +1$. A test set of 1000 observations with no noise was also generated to gauge upon the performance of the methods. Figure 4 shows the decision boundaries for AdaBoost and Random Forests and the corresponding solutions for the QMM algorithm using 500 full-grown trees. The QMM algorithm improves slightly upon the performance of AdaBoost, since it is evident that it does overfit the data set by trying to more closely get a boundary for the noisy points. The random forest solution does not overfit as closely, but its boundary looks more ragged than that of AdaBoost. 

\begin{figure} 
	\subfloat{\includegraphics[width=8.8cm, height=6.3cm]{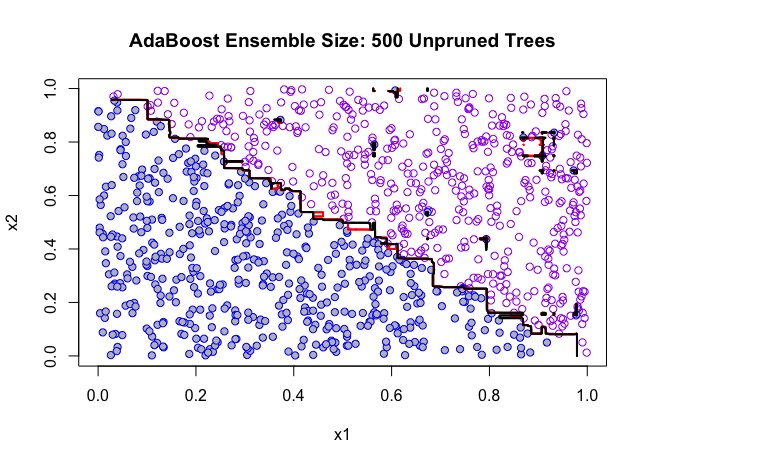}}\
	\subfloat{\includegraphics[width=8.8cm, height=6.3cm]{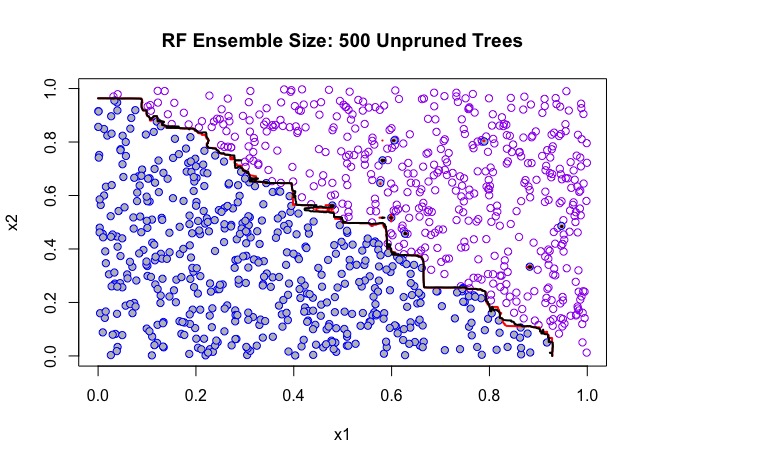}}\
	\caption{Performance of the QMM algorithm with $\upsilon = 0.5$ for AdaBoost and Random Forest using full-grown trees. The test set error for AdaBoost is 0.0280, compared to 0.0240 for the QMM algorithm, pruning the ensemble to only 127 trees. The RF test error rate was  0.0228 compared to QMM test error of 0.0224 selecting only 165 trees.}
	\label{fig:2}
\end{figure}

\begin{figure}[h]
	\centering
	\subfloat{\includegraphics[width=17cm,height=4cm]{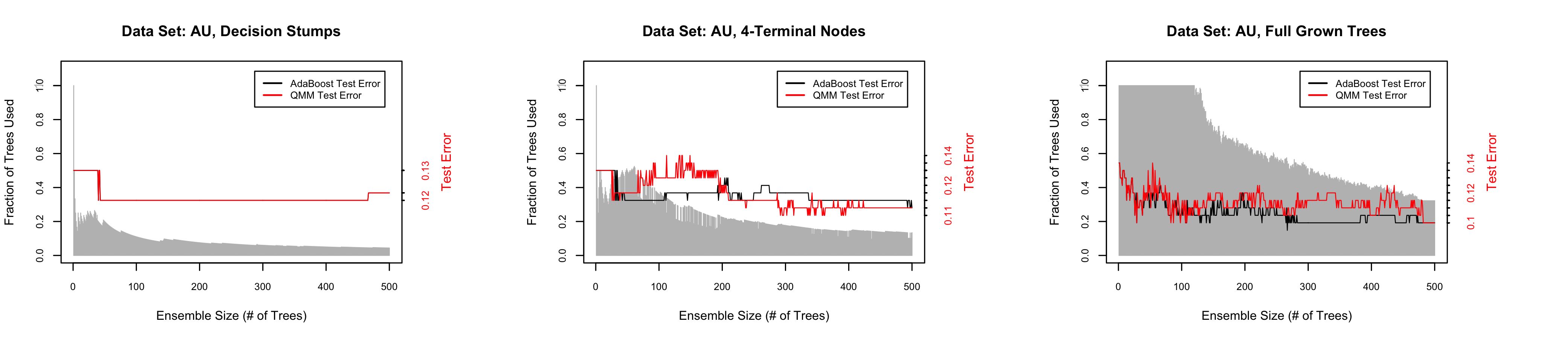}}\
	\subfloat{\includegraphics[width=17cm,height=4cm]{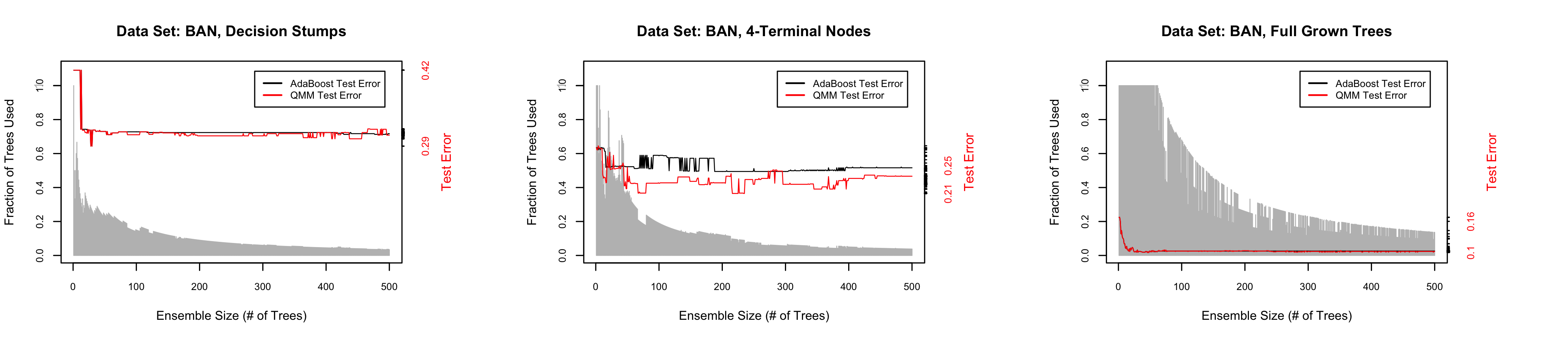}}\
	\subfloat{\includegraphics[width=17cm,height=4cm]{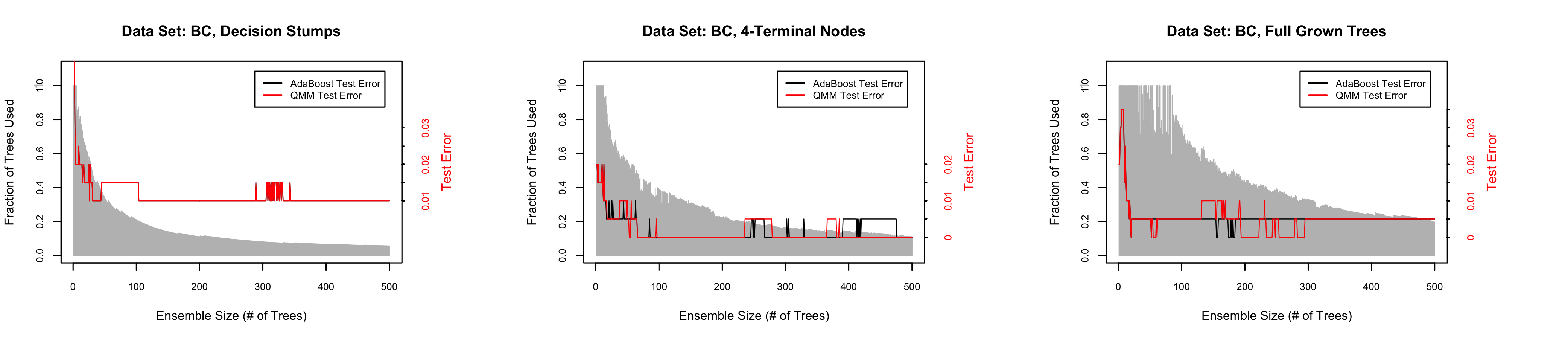}}\
	\subfloat{\includegraphics[width=17cm,height=4cm]{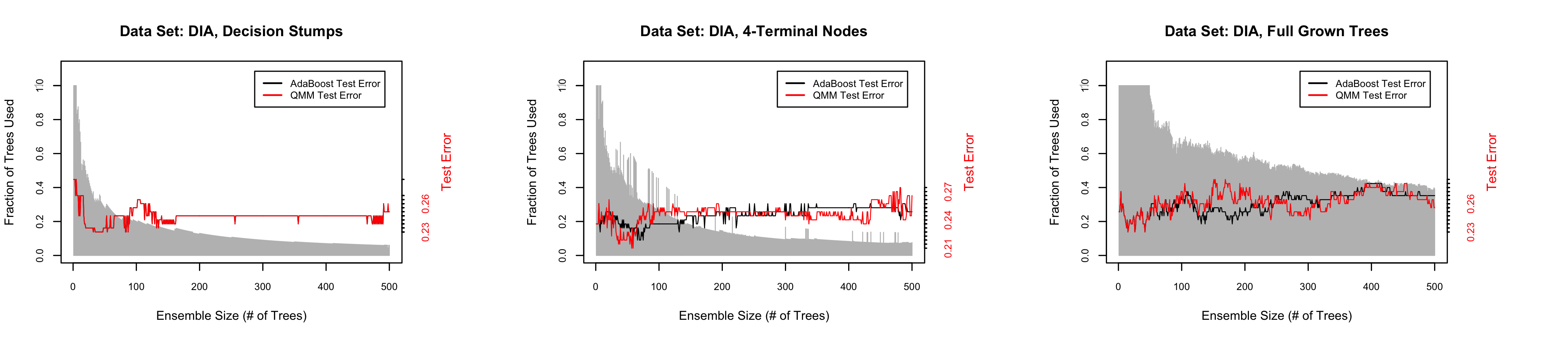}}\
	\subfloat{\includegraphics[width=17cm,height=4cm]{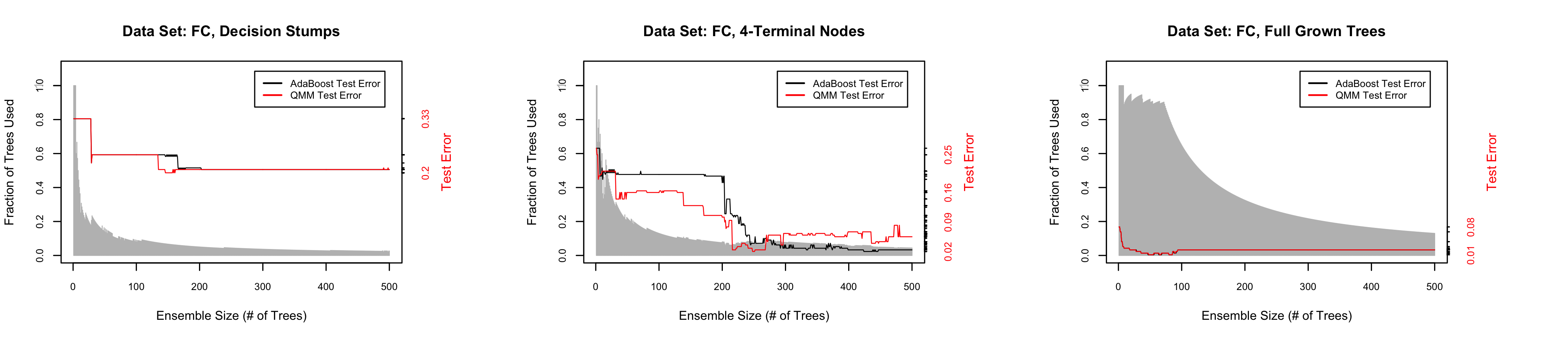}}\
	\caption{Performance of the QMM algorithm vs. AdaBoost for low dimensional data sets using $\upsilon = 0.5$. The shaded grey represents the fraction of weak learners used given the ensemble size, while the red and black curves represent the test set error rates of the QMM algorithm and AdaBoost respectively.}%
	\label{fig:3} 
\end{figure}

\begin{figure}[h]
	\centering
	\subfloat{\includegraphics[width=17cm,height=4cm]{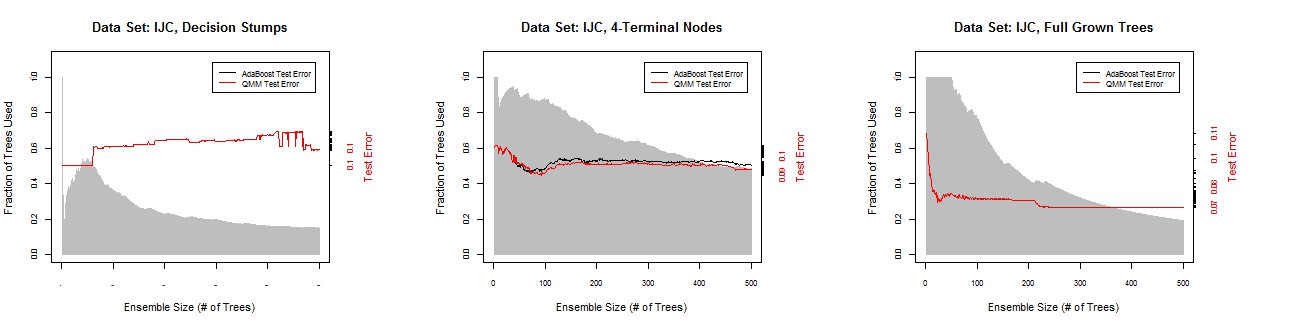}}\
	\subfloat{\includegraphics[width=17cm,height=4cm]{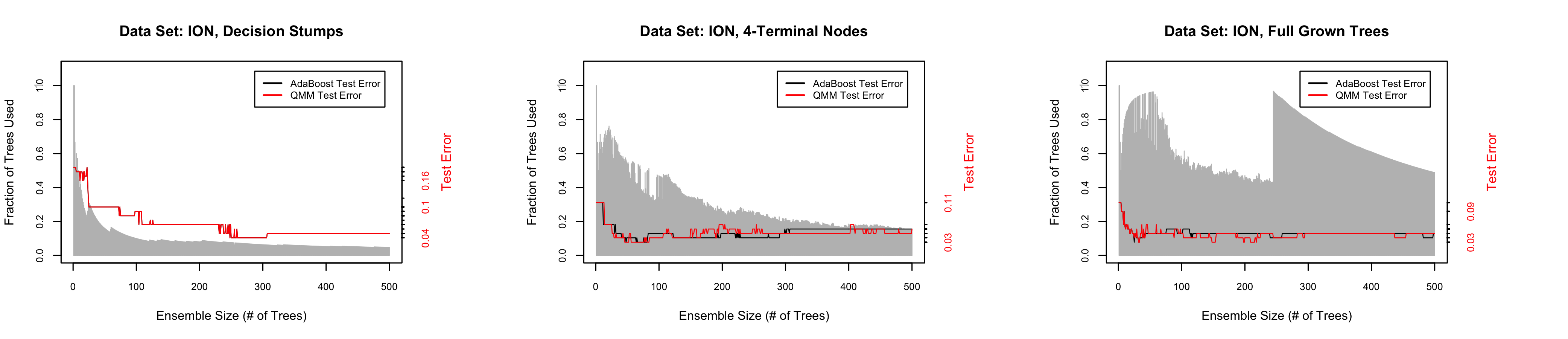}}\
	\subfloat{\includegraphics[width=17cm,height=4cm]{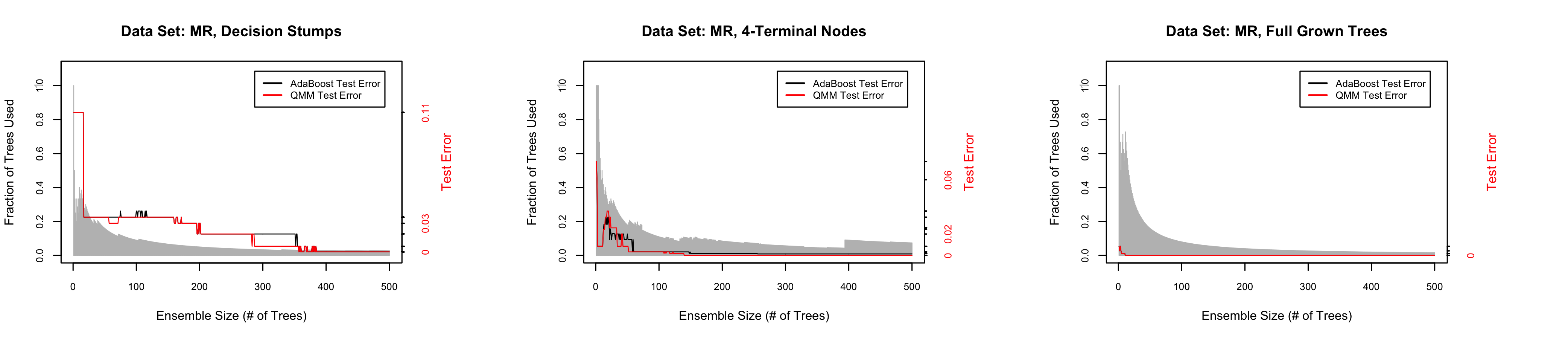}}\
	\subfloat{\includegraphics[width=17cm,height=4cm]{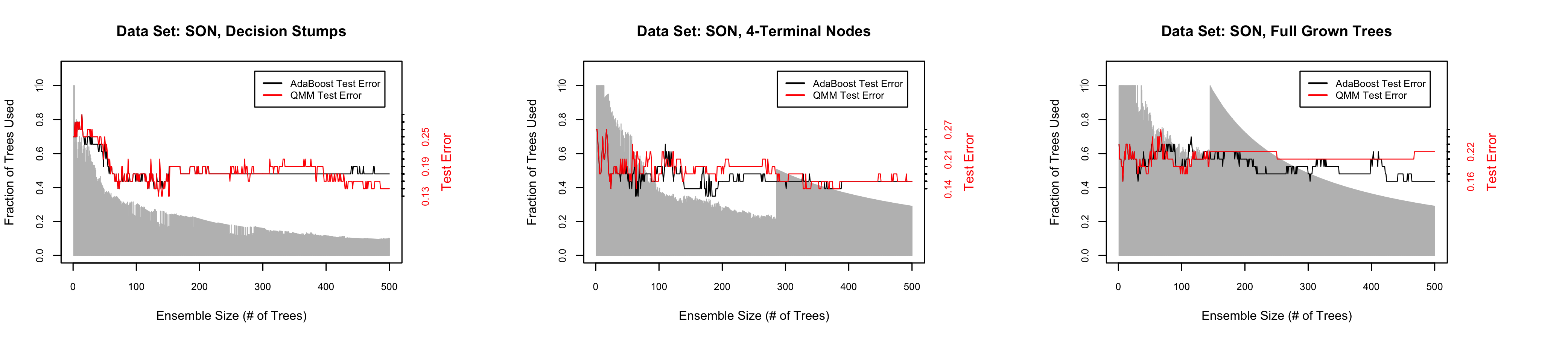}}\
	\subfloat{\includegraphics[width=17cm,height=4cm]{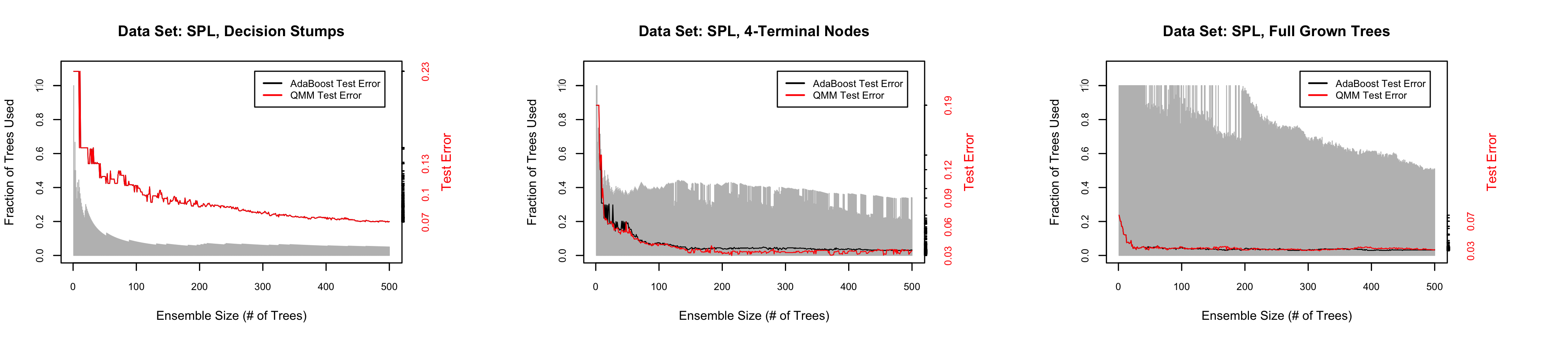}}\
	\caption{Performance of the QMM algorithm vs. AdaBoost for mid dimensional data sets using $\upsilon = 0.5$. The shaded grey represents the fraction of weak learners used given the ensemble size, while the red and black curves represent the test set error rates of the QMM algorithm and AdaBoost respectively.}
	\label{fig:4} 
\end{figure}

\begin{figure}[h]
	\centering
	\subfloat{\includegraphics[width=17cm,height=4cm]{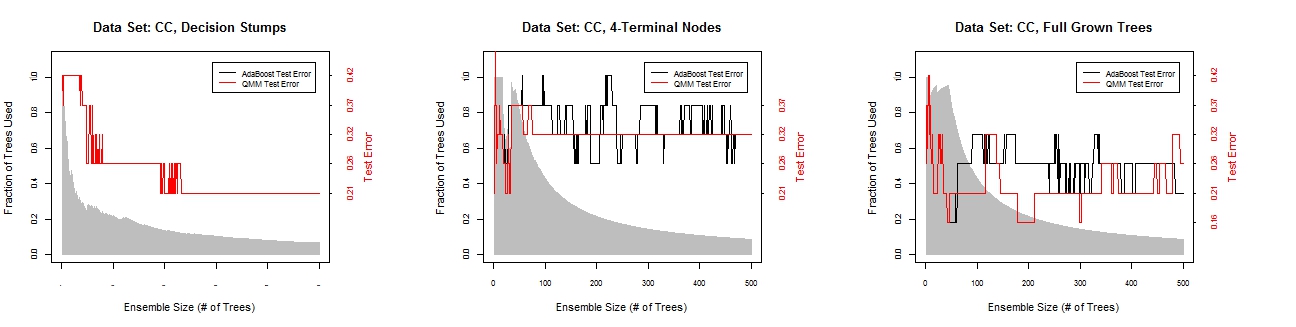}}\
	\subfloat{\includegraphics[width=17cm,height=4cm]{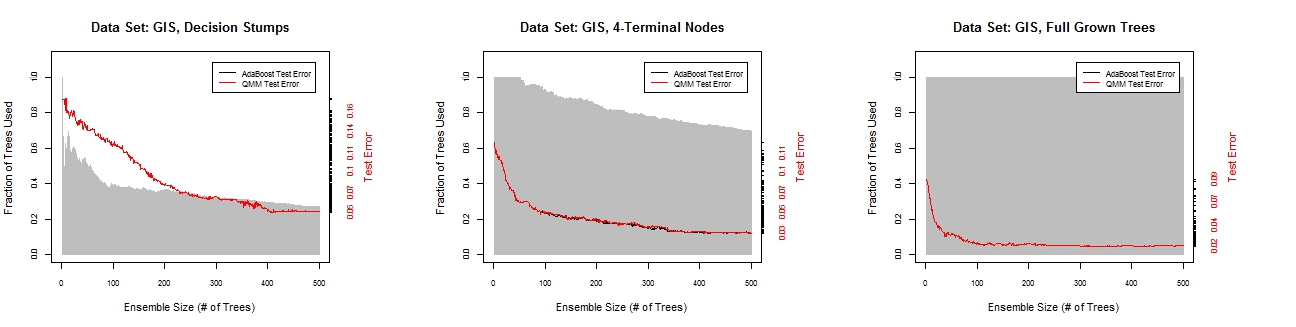}}\
	\subfloat{\includegraphics[width=17cm,height=4cm]{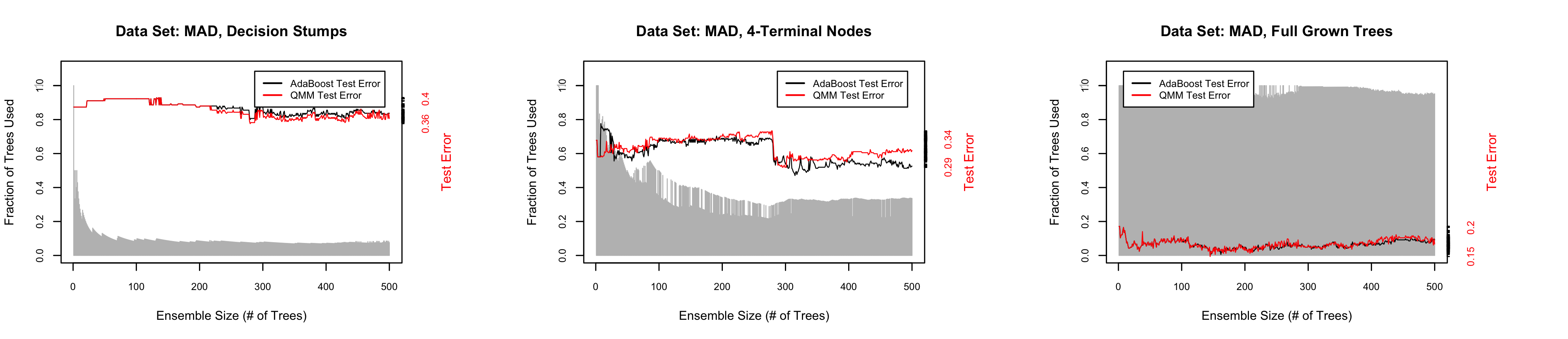}}\

	\caption{Performance of the QMM algorithm vs. AdaBoost for high dimensional data sets using $\upsilon = 0.5$. The shaded grey represents the fraction of weak learners used given the ensemble size, while the red and black curves represent the test set error rates of the QMM algorithm and AdaBoost respectively.}
	\label{fig:5} 
\end{figure}

\begin{figure}[h]
	\centering
	\subfloat{\includegraphics[width=17cm,height=4cm]{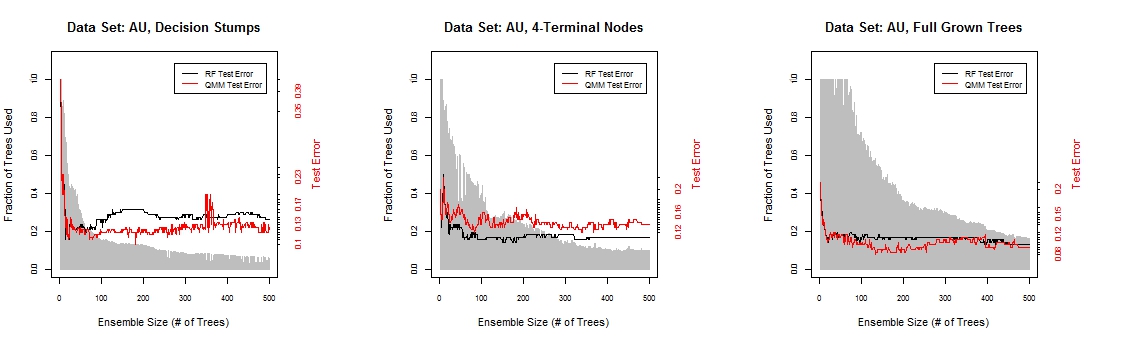}}\
	\subfloat{\includegraphics[width=17cm,height=4cm]{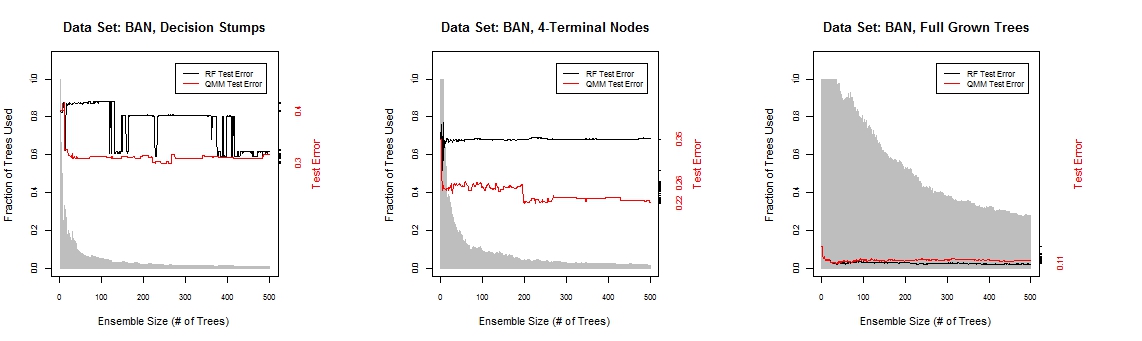}}\
	\subfloat{\includegraphics[width=17cm,height=4cm]{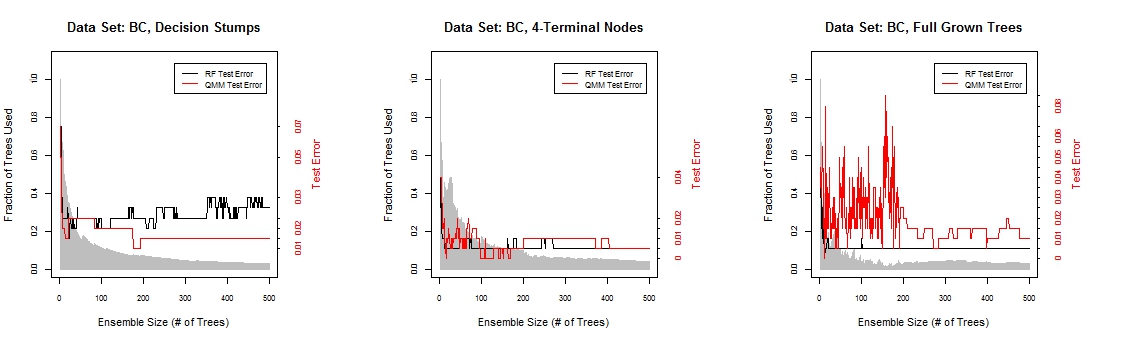}}\
	\subfloat{\includegraphics[width=17cm,height=4cm]{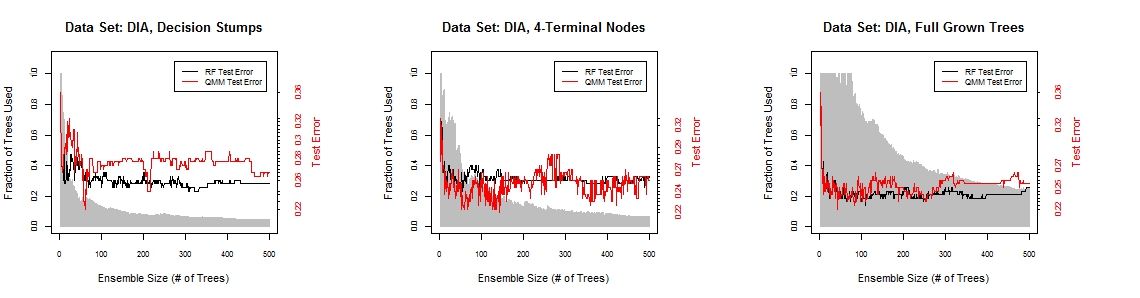}}\
	\subfloat{\includegraphics[width=17cm,height=4cm]{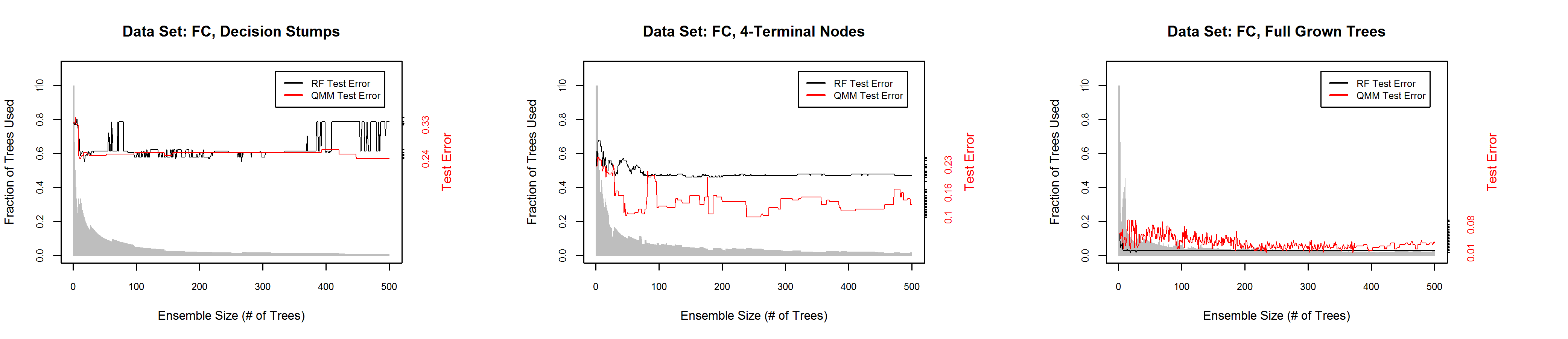}}\
	\caption{Performance of the QMM algorithm vs. random forests for low dimensional data sets using $\upsilon = 0.25$. The shaded grey represents the fraction of weak learners used given the ensemble size, while the red and black curves represent the test set error rates of the QMM algorithm and AdaBoost respectively.}
	\label{fig:6} 
\end{figure}

\begin{figure}[h]
	\centering
	\subfloat{\includegraphics[width=17cm,height=4cm]{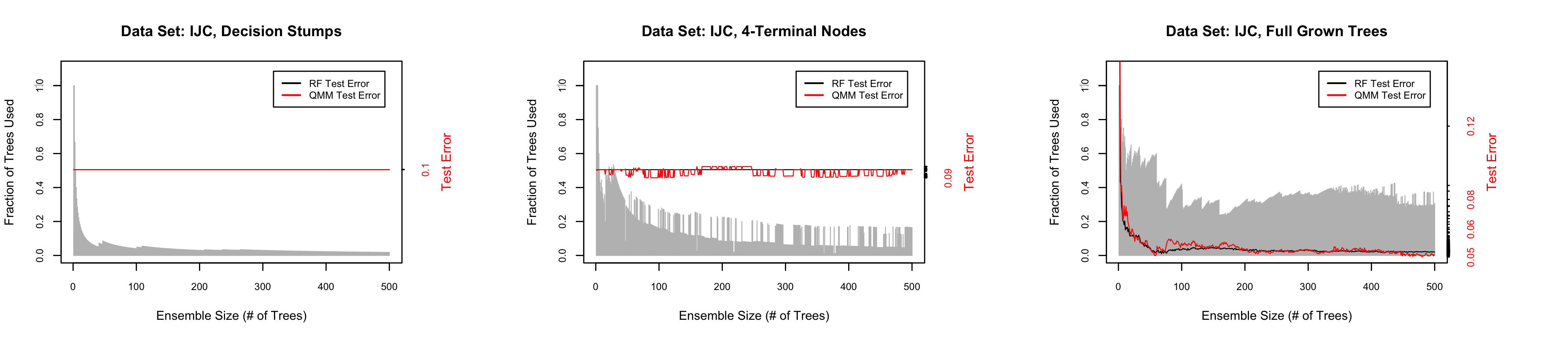}}\
	\subfloat{\includegraphics[width=17cm,height=4cm]{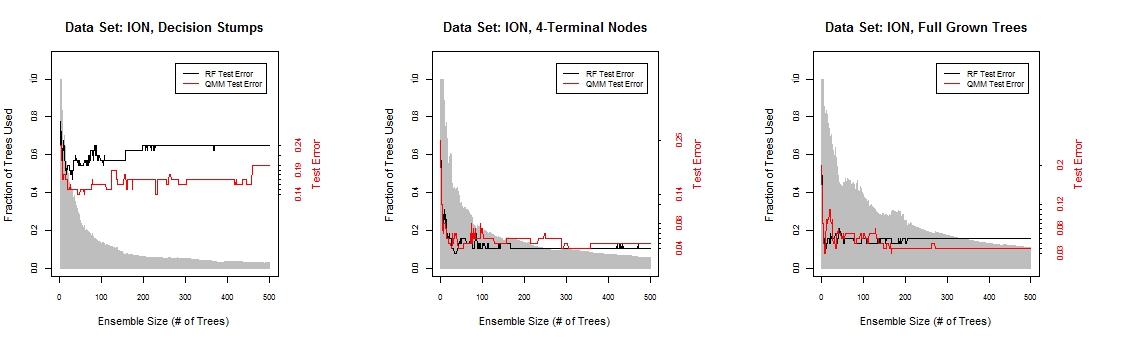}}\
	\subfloat{\includegraphics[width=17cm,height=4cm]{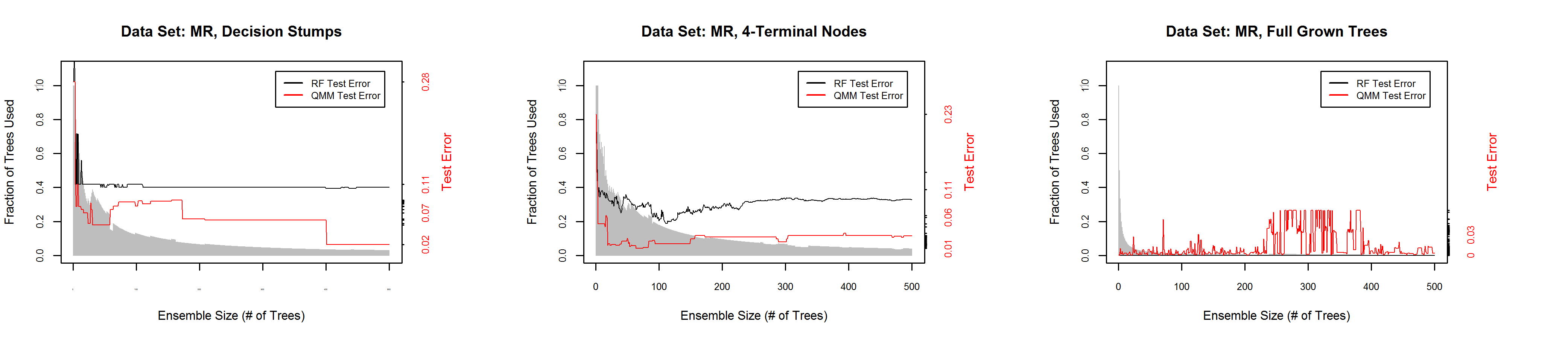}}\
	\subfloat{\includegraphics[width=17cm,height=4cm]{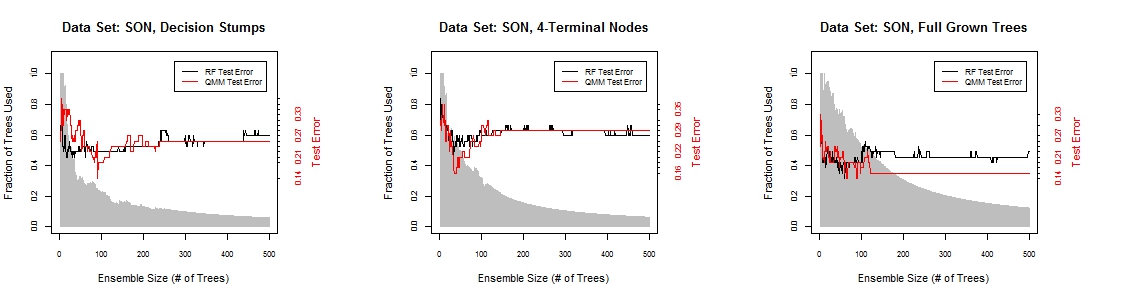}}\
	\subfloat{\includegraphics[width=17cm,height=4cm]{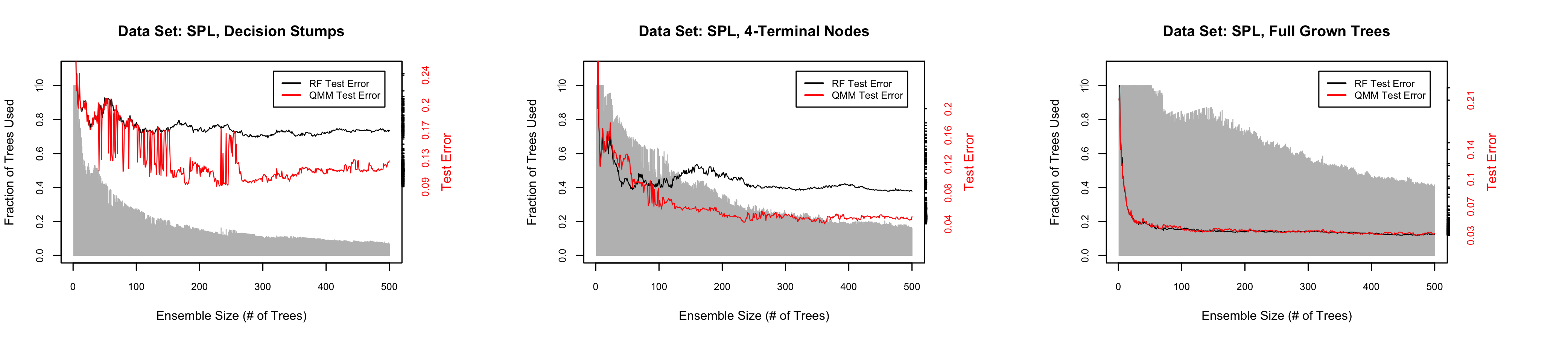}}\
	\caption{Performance of the QMM algorithm vs. random forests for mid dimensional data sets using $\upsilon = 0.25$. The shaded grey represents the fraction of weak learners used given the ensemble size, while the red and black curves represent the test set error rates of the QMM algorithm and AdaBoost respectively.}
	\label{fig:7} 
\end{figure}

\begin{figure}[h]
	\centering
	\subfloat{\includegraphics[width=17cm,height=4cm]{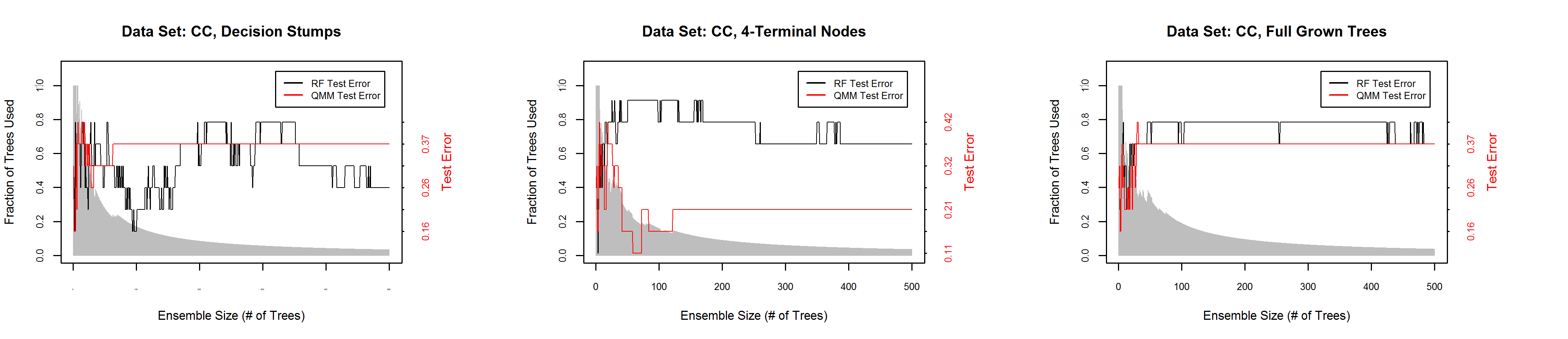}}\
		\subfloat{\includegraphics[width=17cm,height=4cm]{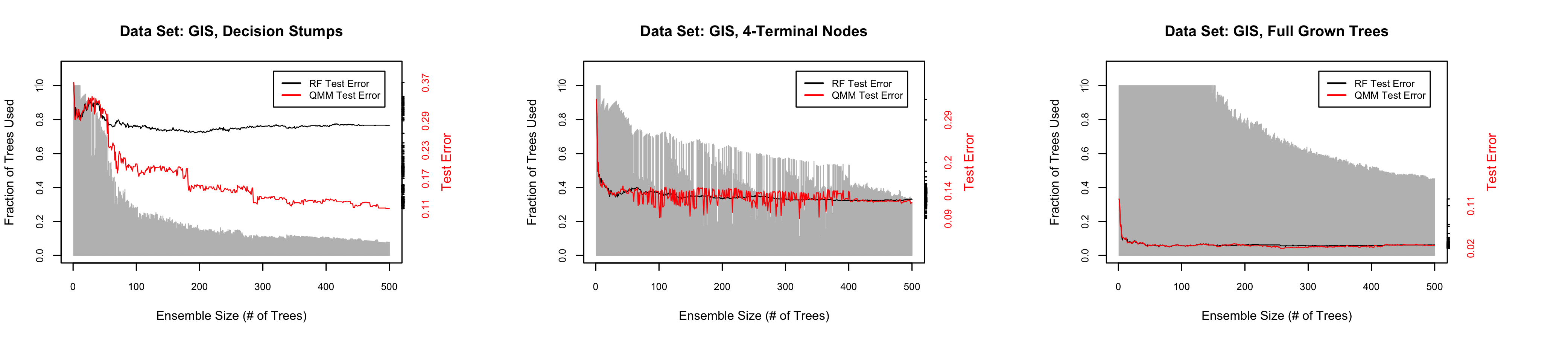}}\
	\subfloat{\includegraphics[width=17cm,height=4cm]{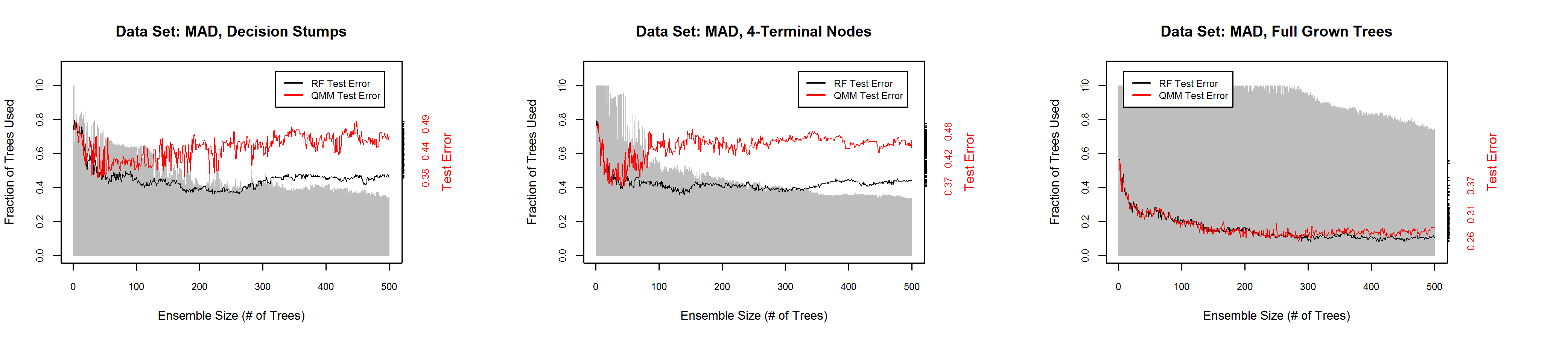}}\
	\caption{Performance of the QMM algorithm vs. random forests for high dimensional data sets using $\upsilon = 0.25$. The shaded grey represents the fraction of weak learners used given the ensemble size, while the red and black curves represent the test set error rates of the QMM algorithm and AdaBoost respectively.}
	\label{fig:8} 
\end{figure}

\subsection{Comparison to Other  Ensemble Pruning Methodologies}

We test how the proposed method compares to two of the leading ensemble pruning techniques: the Diversity Regularized Ensemble Pruning (DREP) method \citep{li2012diversity} and $\kappa$-pruning \citep{margineantu1997pruning}. Both are selection-based methods and considered to be two of the leading ensemble pruning algorithms. One of the main drawbacks of ordered-ensemble pruning methods such as $\kappa$-pruning, is that we need to specify the size of the pruned subensemble and it might not necessarily be the optimal size. Several authors have considered that pruning close to 80\% of a  given ensemble yields the most consistent good results. For that particular reason, we set the $\kappa$-pruning method to prune 80\% of the given ensemble.  We show the test set error rate, the resulting ensemble size, along with a measure of diversity of the method. In this research we define diversity as:

\begin{equation}
\label{eq:25}
\text{div}(f) = 1 - \frac{1}{\sum_{1 \leq i \neq j \leq T}1} \sum_{1 \leq i \neq j \leq T} \text{diff}(h_i, h_j).
\end{equation}
where $\text{diff}(h_i,h_j)= 1/n \sum_{k=1}^n I[h_i(\textbf{x}_k)h_j(\textbf{x}_k)<0]$.
\begin{table}
	\centering
	\caption{Summary for AdaBoost Ensembles. Test Error Average Rank and Average Pruning Rate.}
	\resizebox{\textwidth}{!}{%
		\begin{tabular}{lcccccc}
			\toprule
			& \multicolumn{2}{c}{Decision Stumps} &  \multicolumn{2}{c}{4-Node Trees} & \multicolumn{2}{c}{Full-Grown Trees}    \\
			\midrule
			& Test Error Rank & Pruning Rate &Test Error Rank & Pruning Rate &Test Error Rank & Pruning Rate \\
			\midrule
			AdaBoost &1.31 &  & \textbf{1.46} & & 1.69 &   \\
			QMM & \textbf{1.08} & 0.9229 & \textbf{1.46} & 0.7900 & \textbf{1.54} &  0.6417 \\
			DREP & 3.23 & \textbf{0.9758} & 2.77 & \textbf{0.8943}    & 2.92 & \textbf{0.9518}  \\
			$\kappa$-pruning &3.69 & 0.8000 & 3.46 & 0.8000 & 2.07 &0.8000   \\
			\bottomrule
		\end{tabular}%
		\label{tab:10}%
	}
\end{table}%
\noindent Tables \ref{tab:4}, \ref{tab:5} and \ref{tab:6} summarize the comparison results for AdaBoost ensembles. The simulations suggest that DREP prunes trees more aggresively than both the QMM algorithm and  $\kappa$-pruning, however the QMM algorithm performs better in terms of the test set error in most of the cases, which is not surprising given the primary emphasis of the algorithm to achieve better performance, as opposed to explicitly prune the ensemble.  For low dimensional data sets the QMM algorithm performs very similar to the original AdaBoost ensemble in terms of the test set error, but only using around 12\% of the trees generated by the AdaBoost solution. Table \ref{tab:10} is a breakdown of the average rank of the test error rates for the methods compared, along with the average pruning rate for AdaBoost solutions. In terms of generalization error performance, the average rank of the QMM algorithm is 1.08, 1.46, and 1.54 for decision stumps, 4-node (depth = 2) trees and full-grown trees respective. The QMM algorithm ranks better than DREP and $\kappa$-pruning for decision stumps, 4-node (depth = 2) and full grown trees, and ranks on average better than the original AdaBoost solution for decision stumps and full-grown trees, with a tied performance on 4-node trees. The average pruning rate for the QMM is 92.29\% for decision stumps, 79\% for 4-node trees and 64.17\% for full-grown trees. DREP on average prunes the trees more aggresively than the QMM algorithm with an average pruning rate of 97.58\% for decision stumps, 89.43\% for 4-node trees and 95.18\% for full-grown trees. Tables \ref{tab:7}, \ref{tab:8} and \ref{tab:9} show the comparison results for Random Forest ensembles. For Random Forest solutions, the QMM algorithm ranks best on average in terms of generalization performance for both decision stumps and 4-node (depth = 2) trees, while it ranks second in full-grown trees after the original Random Forest solution. $\kappa$-pruning outperforms DREP only in full-grown trees. In terms of the pruning rate, DREP also obtains on average a higher pruning rate under all decision tree types.

\begin{table}
	\centering
	\caption{Summary for Random Forest Ensembles. Test Error Average Rank and Average Pruning Rate.}
	\resizebox{\textwidth}{!}{%
		\begin{tabular}{lcccccc}
			\toprule
			& \multicolumn{2}{c}{Decision Stumps} &  \multicolumn{2}{c}{4-Node Trees} & \multicolumn{2}{c}{Full-Grown Trees}    \\
			\midrule
			& Test Error Rank & Pruning Rate &Test Error Rank & Pruning Rate &Test Error Rank & Pruning Rate \\
			\midrule
			Random Forest &2.46 & &1.85 & & \textbf{1.84}  & \\
			QMM & \textbf{1.69} & 0.9078 &\textbf{1.62} &  0.7874 & 2.08 & 0.5638   \\
			DREP & 2.08 &\textbf{0.9263} & 2.46&\textbf{0.9245} &  3.15 & \textbf{0.9637}  \\
			$\kappa$-pruning &2.83& 0.8000 & 3.15  & 0.8000 & 2.77 & 0.8000    \\
			\bottomrule
		\end{tabular}%
		\label{tab:11}%
	}
\end{table}%

\section{Conclusions and Future Research}
\label{conc}

Ensembles generally perform strongly in terms of their generalization ability compared to individual classifiers. The proliferation of large scale, high velocity data sets, often containing variables of different data types, creates challenges for the use of ensembles in industry. Recent research effort has concentrated in reducing the size of ensembles, while maintaining their predictive accuracy (pruning the ensemble). In this paper, we propose  a quadratic program formulation that aims to tune the weights of a given ensemble, such that the pairwise correlations of the weak learners  and the variance of the margin instances are minimized, while maximizing the lower percentiles of the margins. The proposed method results in a combined classifier with an average pruning rate of 91.5\% for decision stumps, 78.9\% for 4-node trees and 61.8\% for full-grown trees. The generalization performance of the proposed method (the QMM algorithm) compares favorably to that of the original ensemble and other ensemble pruning methodologies analyzed here. Moreover, the proposed method, as with many ensemble pruning methodologies, appears to offer some improvements over the original ensemble in data sets containing noisy examples. Further directions for this research include to implicitly adhere cost functions in the boosting and random forest ensembles that maximize the lower margin distribution and select the most diverse classifiers. 

\newpage

\begin{table}[htbp]
	\centering
	\caption{Comparison of the QMM Algorithm to Existing Methodologies in Low Dimensional Data Sets (AdaBoost Solutions).}
	\resizebox{\textwidth}{!}{%
		\begin{tabular}{llccccccccc}
			\toprule
			& &\multicolumn{3}{c}{500 Trees (decision stumps)} & \multicolumn{3}{c}{500 Trees (4-terminal nodes)}   & \multicolumn{3}{c}{500 Trees (full-grown)}   \\
			\midrule
			& \multicolumn{1}{l}{\textbf{}} & \multicolumn{1}{c}{Test Error} & \multicolumn{1}{c}{Ensemble Size} & \multicolumn{1}{c}{Diversity} 
			& \multicolumn{1}{c}{Test Error} & \multicolumn{1}{c}{Ensemble Size} & \multicolumn{1}{c}{Diversity} 
			& \multicolumn{1}{c}{Test Error} & \multicolumn{1}{c}{Ensemble Size} & \multicolumn{1}{c}{Diversity} 
			\\
			\midrule
			AU & AdaBoost & \textbf{0.1202} & 500  & 0.51818 & 0.1105 & 500    & 0.52969 &\textbf{0.1009}  & 500 & 0.67624  \\
			& QMM & \textbf{0.1202} & 22  & 0.99923& 0.1105 & 67   &  0.98555  & \textbf{0.1009} & 155   &  0.67624  \\
			& DREP & 0.1346 & 1   & 1.00000 & \textbf{0.1058} & 27   &  0.99873&0.1250& 17  &0.99973  \\
			& $\kappa$-pruning & 0.3894 & 100   &  0.99099& 0.1298 & 100    &  0.98446& 0.1154 & 100   & 0.98560 \\
			\midrule
			BAN & AdaBoost & 0.3086 & 500  & 0.50273 & 0.2502 & 500    & 0.50863&0.1075  & 500 & 0.35591 \\
			& QMM & \textbf{0.3061}  & 17 & 0.99890 & \textbf{0.2357} & 19    &  0.99618  & \textbf{0.1068} & 71   &0.99170 \\
			& DREP & 0.3702 & 7   & 0.99998& 0.2809 & 29  &    0.99835  & 0.1339 & 29   & 0.99779 \\
			& $\kappa$-pruning & 0.3130 & 100   & 0.99636& 0.2778& 100   &   0.98157 & 0.2998 & 100  & 0.96775  \\
			\midrule
			BC & AdatBoost& \textbf{0.0146} & 500  & 0.54122 & \textbf{0.0049} & 500    & 0.58258& \textbf{0.0098}  & 500 & 0.86199 \\
			& QMM & \textbf{0.0146}  & 28 & 0.99899& \textbf{0.0049} & 55    &  0.98027 &  \textbf{0.0098} & 98  &0.94103 \\
			& DREP & 0.0537 & 5   & 0.99999& 0.0098 & 37  &  0.99814 & 0.0195 & 11   & 0.99996  \\
			& $\kappa$-pruning & 0.0342 & 100   & 0.99151 &0.0146& 100   &   0.98788 & 0.0146 & 100  & 0.99299  \\
			\midrule
			DIA & AdaBoost & \textbf{0.2468}& 500  & 0.53804 & 0.2468 & 500    &0.54749 &0.2641  & 500 & 0.60059 \\
			& QMM & \textbf{0.2468}  & 30 & 0.99845 & 0.2641 & 38    &  0.99175 & \textbf{0.2511} & 211   &0.65181 \\
			& DREP &  0.2641 & 5  & 0.99996 & \textbf{0.2337} & 26  &   0.99890   & 0.2641 &38   &0.99802 \\
			& $\kappa$-pruning & 0.3550 & 100   & 0.98450 & 0.2597& 100   &   0.98312& \textbf{0.2511} & 100  & 0.98399 \\
			\midrule
			FC & AdaBoost & \textbf{0.2124} & 500  & 0.54855 & \textbf{0.0232} & 500    & 0.57073&\textbf{0.0232}  & 500 &0.98695 \\
			& QMM & \textbf{0.2124}  & 13 & 0.99971& 0.0541 & 22    &  0.99441& \textbf{0.0232}  & 65   &0.99906 \\
			& DREP & 0.3320 & 3   &0.99999& 0.2625 & 23  &   0.99925  & \textbf{0.0232} & 3   &1.00000 \\
			& $\kappa$-pruning & 0.4285 & 100   & 0.99382& 0.1737& 100   &   0.98305 & \textbf{0.0232}  & 100  & 0.99810 \\
			\bottomrule
		\end{tabular}%
		\label{tab:4}%
	}
\end{table}%

\begin{table}[htbp]
	\centering
	\caption{Comparison of the QMM Algorithm to Existing Methodologies in Mid Dimensional Data Sets (AdaBoost Solutions). }
	\resizebox{\textwidth}{!}{%
		\begin{tabular}{llccccccccc}
			\toprule
			& &\multicolumn{3}{c}{500 Trees (decision stumps)} & \multicolumn{3}{c}{500 Trees (4-terminal nodes)}   & \multicolumn{3}{c}{500 Trees (full-grown)}   \\
			\midrule
			& \multicolumn{1}{l}{\textbf{}} & \multicolumn{1}{c}{Test Error} & \multicolumn{1}{c}{Ensemble Size} & \multicolumn{1}{c}{Diversity} 
			& \multicolumn{1}{c}{Test Error} & \multicolumn{1}{c}{Ensemble Size} & \multicolumn{1}{c}{Diversity} 
			& \multicolumn{1}{c}{Test Error} & \multicolumn{1}{c}{Ensemble Size} & \multicolumn{1}{c}{Diversity} 
			\\
			\midrule
			IJC & AdaBoost & 0.1040 & 500  & 0.52748& 0.0959 & 500    & 0.54184 &0.0740 & 500  & 0.87387  \\
			& QMM & 0.1040 & 75  & 0.99047  & \textbf{0.0935} & 242   &  0.89703 & 0.0740 & 96  &  0.99209\\
			& DREP & \textbf{0.0959} & 2   & 1.00000 & 0.1107 & 148   &   0.96062& 0.0972 & 15   & 0.98352  \\
			& $\kappa$-pruning & 0.3393 & 100   &  0.98360 & 0.1162 & 100    &  0.98295 &  \textbf{0.0699} & 100   &  0.99640 \\
			\midrule
			ION & AdaBoost & \textbf{0.0472} & 500  & 0.52661 & \textbf{0.0566} & 500    & 0.56099 &0.0472  & 500 & 0.76899 \\
			& QMM & \textbf{0.0472}  & 24 & 0.99921& \textbf{0.0566} & 76    &  0.95103  & 0.0472 & 244   &0.94753\\
			& DREP & 0.1132 & 20   & 0.99927&0.0660 & 7 &  0.99994  & 0.1132 & 6   & 0.99999 \\
			& $\kappa$-pruning & 0.1981 & 100   & 0.99087& 0.3585 & 100   &   0.98245 & \textbf{0.0377} & 100  & 0.98810  \\
			\midrule
			MR & AdatBoost& \textbf{0.0029} & 500  & 0.54464 & 0.0012 & 500    &  0.58310 & \textbf{0.0000}  & 500 & 0.99985 \\
			& QMM & \textbf{0.0029}  & 14 & 0.99955& \textbf{0.0000} & 37    &  0.99778 &  \textbf{0.0000} & 8  &1.00000 \\
			& DREP & 0.1148 & 1   & 1.00000& 0.0066 & 10  &   0.99988  &  \textbf{0.0000} & 2  & 1.00000 \\
			& $\kappa$-pruning & 0.0849 & 100   & 0.98607 &0.0221 & 100   &   0.99393 &  \textbf{0.0000}& 100  & 0.99997  \\
			\midrule
			SON & AdaBoost &0.1746& 500  & 0.51665 &\textbf{0.1587} & 500 &0.54364    & \textbf{0.1587}& 500  & 0.59817 \\
			& QMM & \textbf{0.1429}  & 51 & 0.99088 & \textbf{0.1587}& 145    &  0.96215  & 0.2222  & 145 & 0.96701 \\
			& DREP &  0.2381 & 68  & 0.99124&0.2222& 26  &   0.99884  & 0.2381 & 10  & 0.99988 \\
			& $\kappa$-pruning & 0.3175  & 100   & 0.98168& 0.1746& 100   &   0.981742 & 0.1905  & 100   & 0.98273  \\
			\midrule
			SPL & AdaBoost & \textbf{0.0662} & 500  & 0.52568& 0.0359 & 500    &0.55834 &\textbf{0.0359}  & 500 &0.76731 \\
			& QMM & \textbf{0.0662}  & 25 & 0.99883& \textbf{0.0349} & 170    &  0.94612 & 0.0363  & 255   &0.76731 \\
			& DREP & 0.1651 & 42  &0.99674& 0.0446 & 103  &   0.98297 & 0.0556 & 4   & 1.00000 \\
			& $\kappa$-pruning & 0.3402 & 100   & 0.98732& 0.1582& 100   &   0.98480 & 0.0363  & 100  & 0.99010  \\
			\bottomrule
		\end{tabular}%
		\label{tab:5}%
	}
\end{table}%

\begin{table}[htbp]
	\centering
	\caption{Comparison of the QMM Algorithm to Existing Methodologies in High Dimensional Data Sets (AdaBoost Solutions).}
	\resizebox{\textwidth}{!}{%
		\begin{tabular}{llccccccccc}
			\toprule
			& &\multicolumn{3}{c}{500 Trees (decision stumps)} & \multicolumn{3}{c}{500 Trees (4-terminal nodes)}   & \multicolumn{3}{c}{500 Trees (full-grown)}   \\
			\midrule
			& \multicolumn{1}{l}{\textbf{}} & \multicolumn{1}{c}{Test Error} & \multicolumn{1}{c}{Ensemble Size} & \multicolumn{1}{c}{Diversity} 
			& \multicolumn{1}{c}{Test Error} & \multicolumn{1}{c}{Ensemble Size} & \multicolumn{1}{c}{Diversity} 
			& \multicolumn{1}{c}{Test Error} & \multicolumn{1}{c}{Ensemble Size} & \multicolumn{1}{c}{Diversity} 
			\\
			\midrule
			CC & AdaBoost &\textbf{0.2105}& 500  & 0.62274 & \textbf{0.3684} & 500    & 0.00000 &0.3684& 500  & 0.59920   \\
			& QMM & \textbf{0.2105}& 29  & 0.99814& \textbf{0.3684} & 1   &  1.00000 & \textbf{0.3158} & 42  & 0.99770 \\
			& DREP & 0.4210 & 1   & 1.00000& \textbf{0.3684}& 1   &  1.00000& 0.4210 & 2   & 0.99999  \\
			& $\kappa$-pruning & 0.2631& 100   &  0.98365& \textbf{0.3684} & 100    &  0.96032 & \textbf{0.3158}& 100   &  0.99132\\
			\midrule
			GIS & AdaBoost & \textbf{0.0540}& 500  & 0.94414 & \textbf{0.0320} & 500    & 0.93792&\textbf{0.0200}  & 500 & 0.52574 \\
			& QMM & \textbf{0.0540}  & 135 & 0.99469&  \textbf{0.0320} & 325    &  0.95839 & \textbf{0.0200}  & 465  & 0.52574 \\
			& DREP & 0.1690& 1   & 1.00000& 0.0390 & 177  &   0.98646  & 0.0210& 177   & 0.93672\\
			& $\kappa$-pruning & 0.1400 & 100   & 0.99093& 0.0720 & 100   &   0.99364 & 0.0210 & 100  & 0.98647  \\
			\midrule
			MAD & AdaBoost& 0.3783 & 500  & 0.50401 & \textbf{0.2933} & 500    &  0.50689& 0.1783 & 500 & 0.52150  \\
			& QMM & \textbf{0.3700}  & 38 & 0.99660 & 0.3200 & 168    &  0.94464  &  \textbf{0.1683} & 474   & 0.52150 \\
			& DREP & 0.3883 & 1  & 1.00000& 0.3450& 87 &   0.99998 & 0.1967 & 75  & 0.98963 \\
			& $\kappa$-pruning & 0.4400 & 100   &0.98085 &0.3266& 100   &  0.98296& 0.2250 & 100  & 0.98056  \\
			\bottomrule
		\end{tabular}%
		\label{tab:6}%
	}
\end{table}%

\begin{table}[htbp]
	\centering
	\caption{Comparison of the QMM Algorithm to Existing Methodologies in Low Dimensional Data Sets (Random Forest Solutions).}
	\resizebox{\textwidth}{!}{%
		\begin{tabular}{llccccccccc}
			\toprule
			& &\multicolumn{3}{c}{500 Trees (decision stumps)} & \multicolumn{3}{c}{500 Trees (4-terminal nodes)}   & \multicolumn{3}{c}{500 Trees (full-grown)}   \\
			\midrule
			& \multicolumn{1}{l}{\textbf{}} & \multicolumn{1}{c}{Test Error} & \multicolumn{1}{c}{Ensemble Size} & \multicolumn{1}{c}{Diversity} 
			& \multicolumn{1}{c}{Test Error} & \multicolumn{1}{c}{Ensemble Size} & \multicolumn{1}{c}{Diversity} 
			& \multicolumn{1}{c}{Test Error} & \multicolumn{1}{c}{Ensemble Size} & \multicolumn{1}{c}{Diversity} 
			\\
			\midrule
			AU & RF &0.1442 & 500  & 0.68613&\textbf{0.1105} & 500  & 0.73518 &0.0961 & 500  &  0.75578 \\
			& QMM & \textbf{0.1250} & 32  & 0.99819 &0.1442 & 102 &  0.96848& \textbf{0.0913} & 282  & 0.77948\\
			& DREP & \textbf{0.1250} & 51   &0.99692&0.1106 & 52   & 0.99734&  0.1250 & 18   & 0.99968 \\
			& $\kappa$-pruning & 0.2500 & 100   &  0.98616&0.1538  & 100   &  0.98897 & 0.0961 & 100   &  0.98832\\
			\midrule
			BAN & RF &0.3243& 500  & 0.69109 & 0.3494 & 500  & 0.76218 &\textbf{0.1068}& 500  & 0.87735 \\
			& QMM & \textbf{0.3193} & 4 & 0.99997 & \textbf{0.2099} & 14 & 0.99962  & 0.1081 & 470 & 0.87735  \\
			& DREP &0.4192 & 11 &0.99985& 0.3620 &3 &0.99999 &  0.1087 & 45  &0.99903 \\
			& $\kappa$-pruning & 0.4173  & 100   &0.99643& 0.4028  & 100   & 0.99145 & 0.1119  & 100   & 0.99460  \\
			\midrule
			BC & RF& 0.0292 & 500  & 0.90661& \textbf{0.0098} & 500  & 0.93904 & \textbf{0.0098} & 500  & 0.94265 \\
			& QMM & \textbf{0.0146}  & 14 & 0.99988&\textbf{0.0098} & 34& 0.99964&  0.0146  & 15 & 0.99993 \\
			& DREP & 0.0292 & 9  & 0.99997&\textbf{0.0098} & 11   & 0.99998 & 0.0341 & 4   & 1.00000 \\
			& $\kappa$-pruning & 0.0376 & 100   &0.99615 & 0.0195 & 100   &0.99613&  0.0146 & 100   &0.99697  \\
			\midrule
			DIA & RF & \textbf{0.2510} & 500  & 0.67500  & 0.2640 & 500  &  0.71681 &\textbf{0.2467} & 500  & 0.67500 \\
			& QMM & 0.2641  & 31 &0.67641 & \textbf{0.2554}  & 126 & 0.94086 & 0.2511  & 300 & 0.67632\\
			& DREP &  0.2554 & 58  & 0.99969  & \textbf{0.2554} & 51  & 0.99695 & 0.2640 & 16  & 0.99969 \\
			& $\kappa$-pruning & 0.3246& 100   & 0.98610 & 0.2727& 100   & 0.99068 &0.2554& 100   & 0.98610  \\
			\midrule
			FC & RF & 0.3359 & 500  & 0.71945& 0.2007 & 500  & 0.78693 &\textbf{0.0116} & 500  & 0.91949 \\
			& QMM & \textbf{0.2432}  & 4 & 0.99998& \textbf{0.1737}  &10 &0.99985 & 0.0309  &13 & 0.99992 \\
			& DREP & 0.2780  & 4   &0.99589& 0.2896  & 9 &0.99995& 0.0232  & 5   &0.99999\\
			& $\kappa$-pruning & 0.3243 & 100   &  0.99711 & 0.2278  & 100   & 0.99766 & 0.0154  & 100   & 0.99535 \\
			\bottomrule
		\end{tabular}%
		\label{tab:7}%
	}
\end{table}%

\begin{table}[htbp]
	\centering
	\caption{Comparison of the QMM Algorithm to Existing Methodologies in Mid Dimensional Data Sets (Random Forest Solutions).}
	\resizebox{\textwidth}{!}{%
		\begin{tabular}{llccccccccc}
			\toprule
			& &\multicolumn{3}{c}{500 Trees (decision stumps)} & \multicolumn{3}{c}{500 Trees (4-terminal nodes)}   & \multicolumn{3}{c}{500 Trees (full-grown)}   \\
			\midrule
			& \multicolumn{1}{l}{\textbf{}} & \multicolumn{1}{c}{Test Error} & \multicolumn{1}{c}{Ensemble Size} & \multicolumn{1}{c}{Diversity} 
			& \multicolumn{1}{c}{Test Error} & \multicolumn{1}{c}{Ensemble Size} & \multicolumn{1}{c}{Diversity} 
			& \multicolumn{1}{c}{Test Error} & \multicolumn{1}{c}{Ensemble Size} & \multicolumn{1}{c}{Diversity} 
			\\
			\midrule
			IJC & RF &\textbf{0.0959} & 500  & 1.00000 & \textbf{0.0959} & 500  & 0.99130  &0.0520 & 500  & 0.32555   \\
			& QMM & \textbf{0.0959} & 12  & 1.00000&\textbf{0.0959} & 70  & 0.99968  & 0.0522 & 367  & 0.328200 \\
			& DREP & \textbf{0.0959} & 3   & 1.00000& \textbf{0.0959} & 4  & 1.00000& 0.0555 & 59   & 0.99091  \\
			& $\kappa$-pruning & \textbf{0.0959} & 100   &  1.00000&  \textbf{0.0959} & 100   &  0.99867 &  \textbf{0.0466} & 100   &  0.97544\\
			\midrule
			ION & RF &  0.2358  & 500  & 0.50635 & \textbf{0.0377}  & 500  & 0.77674& 0.0566  & 500  & 0.81124 \\
			& QMM & \textbf{0.1887} & 32 & 0.99579& 0.0472 & 96 & 0.97949 &  \textbf{0.0377} & 111 & 0.97477 \\
			& DREP &  0.2358& 2   & 0.99999& 0.0755& 13   & 0.99987  &  0.0471& 7   & 0.99997  \\
			& $\kappa$-pruning &  0.2170& 100   & 0.98236& 0.1132 & 100   & 0.99065 &  0.0471 & 100   & 0.99069  \\
			\midrule
			MR & RF& 0.1095 & 500  & 0.70385 & 0.0890& 500    & 0.77811& \textbf{0.0000}  & 500 & 0.99164\\
			& QMM & \textbf{0.0689} &31 & 0.99947& \textbf{0.0422}& 68   &0.99181 &  0.0037 & 1   &1.00000 \\
			& DREP & 0.1140& 3   & 0.99999& 0.0972 & 11  & 0.99993 & 0.0008& 2  &1.00000 \\
			& $\kappa$-pruning & 0.2506& 100   & 0.98546 &0.1058 & 100  & 0.98800 & \textbf{0.0000}  & 100  & 0.99928 \\
			\midrule
			SON & RF & 0.2698 & 500  & 0.85220 & 0.2698 & 500  & 0.79122&0.2222 & 500  & 0.79578 \\
			& QMM & 0.2539  & 40 & 0.99893 &  0.2857  & 82 & 0.98355 &  \textbf{0.1587}  & 99 & 0.98402\\
			& DREP &  \textbf{0.2063} & 35 &0.99919&  0.2857 & 13  &0.99986  &  0.1905 & 6  &0.99998 \\
			& $\kappa$-pruning & 0.1905 & 100   & 0.99469& \textbf{0.2063} & 100   & 0.99513 & 0.2381 & 100   & 0.99513  \\
			\midrule
			SPL & RF & 0.1678 & 500  &0.84542  & 0.0869 & 500  & 0.79140 &0.0285 & 500  & 0.62272 \\
			& QMM & \textbf{0.1264} & 41 &  0.99926& \textbf{0.0483}  & 172 & 0.96441 & \textbf{0.0280}  & 356 & 0.62272 \\
			& DREP & 0.1651 & 91   &0.99556& 0.0993 & 119   &0.98794  & 0.0662 & 13   &0.99976 \\
			& $\kappa$-pruning & 0.3457 & 100   & 0.99035& 0.1660 & 100   & 0.99334 & 0.0303 & 100   & 0.98926 \\
			\bottomrule
		\end{tabular}%
		\label{tab:8}%
	}
\end{table}%

\begin{table}[htbp]
	\centering
	\caption{Comparison of the QMM Algorithm to Existing Methodologies in High Dimensional Data Sets (Random Forest Solutions).}
	\resizebox{\textwidth}{!}{%
		\begin{tabular}{llccccccccc}
			\toprule
			& &\multicolumn{3}{c}{500 Trees (decision stumps)} & \multicolumn{3}{c}{500 Trees (4-terminal nodes)}   & \multicolumn{3}{c}{500 Trees (full-grown)}   \\
			\midrule
			& \multicolumn{1}{l}{\textbf{}} & \multicolumn{1}{c}{Test Error} & \multicolumn{1}{c}{Ensemble Size} & \multicolumn{1}{c}{Diversity} 
			& \multicolumn{1}{c}{Test Error} & \multicolumn{1}{c}{Ensemble Size} & \multicolumn{1}{c}{Diversity} 
			& \multicolumn{1}{c}{Test Error} & \multicolumn{1}{c}{Ensemble Size} & \multicolumn{1}{c}{Diversity} 
			\\
			\midrule
			CC & AdaBoost & 0.2631& 500  &0.72614 & 0.3684 & 500  & 0.69650 &0.3684& 500  & 0.70117   \\
			& QMM & 0.3158 & 27  & 0.99822&\textbf{0.1579} & 33  &0.99808  & 0.3684 & 29  & 0.99843\\
			& DREP & 0.3158 & 8   & 0.99995&  0.2105 & 2   & 0.99999&  \textbf{0.2105} & 4   & 0.99997 \\
			& $\kappa$-pruning & \textbf{0.1579} & 100   &  0.99558& 0.2632 & 100   &  0.99394 & 0.2631 & 100   &  0.99366 \\
			\midrule
			GIS & AdaBoost & 0.2810& 500  & 0.71720& 0.1260 & 500    & 0.74891&0.0320 & 500 & 0.84613 \\
			& QMM & \textbf{0.1060}  & 109 & 0.96065& \textbf{0.1040} & 285    &   0.81060  & \textbf{0.0310} & 358  &0.84613 \\
			& DREP & 0.2770& 13   & 0.99984& 0.1320& 190 &   0.96449& 0.0340 & 25   & 0.99963 \\
			& $\kappa$-pruning & 0.3410& 100   & 0.98808& 0.1750& 100   &   0.98848& 0.0340 & 100  & 0.99314 \\
			\midrule
			MAD & AdatBoost& 0.3883 & 500  & 0.95258&\textbf{0.3800} & 500  & 0.92124& \textbf{0.2650} & 500  & 0.96271 \\
			& QMM & 0.4550 & 222 & 0.99895&0.4417& 290 & 0.96039 &  0.2967 & 434 & 0.96296\\
			& DREP & \textbf{0.3683} & 191   & 0.99221& 0.4167 & 13  & 0.99991 & 0.3250 & 32   & 0.99984 \\
			& $\kappa$-pruning & 0.3850 & 100   & 0.99071 &\textbf{0.3800} & 100   &  0.99353& 0.3267 & 100   & 0.99922 \\
			\bottomrule
		\end{tabular}%
		\label{tab:9}%
	}
\end{table}%

\newpage

\appendix

\bibliographystyle{plainnat}
\end{document}